\pdfoutput=1
\documentclass[11pt]{article}

\usepackage[]{EMNLP2023}

\usepackage{times}
\usepackage{latexsym}

\usepackage[T1]{fontenc}
\usepackage[utf8]{inputenc}

\usepackage{microtype}

\usepackage{inconsolata}
\usepackage{booktabs}
\usepackage{graphicx}
\usepackage{amssymb}
\usepackage{adjustbox}
\usepackage{multirow}
\usepackage{multicol}

\title{Machine Translation for Nko: Resources and Baseline System}
\title{Machine Translation for Nko\\Tools, Corpora and Baseline Results}
\title{Machine Translation for Nko: Tools, Corpora and Baseline Results}

\author{First Author \\
  Affiliation / Address line 1 \\
  Affiliation / Address line 2 \\
  Affiliation / Address line 3 \\
  \texttt{email@domain} \\\And
  Second Author \\
  Affiliation / Address line 1 \\
  Affiliation / Address line 2 \\
  Affiliation / Address line 3 \\
  \texttt{email@domain} \\}

\author{Moussa Koulako B. Doumbouya \\
  Stanford Univer / Address line 1 \\
  Affiliation / Address line 2 \\
  Affiliation / Address line 3 \\
  \texttt{email@domain} \\\And
  Second Author \\
  Affiliation / Address line 1 \\
  Affiliation / Address line 2 \\
  Affiliation / Address line 3 \\
  \texttt{email@domain} \\}
\author{Moussa Koulako Bala Doumbouya \thanks{moussa@cs.stanford.edu} $^{\mathcal{S},\mathcal{F}}$
    {\bf Baba Mamadi Diané $^\mathcal{N}$}
    {\bf Solo Farabado Cissé $^\mathcal{N}$}  \\
    {\bf Djibrila Diané $^\mathcal{N}$} 
    {\bf Abdoulaye Sow $^\mathcal{F}$} 
    {\bf Séré Moussa Doumbouya $^\mathcal{F}$} \\
    {\bf Daouda Bangoura $^\mathcal{F}$} 
    {\bf Fodé Moriba Bayo $^\mathcal{F}$} 
    {\bf Ibrahima Sory 2. Condé$^\mathcal{K}$} \\
    {\bf Kalo Mory Diané $^\mathcal{N}$} 
    {\bf Chris Piech $^\mathcal{S}$}
    {\bf Christopher Manning $^\mathcal{S}$}\\\\
    $^\mathcal{S}$ Computer Science Department, Stanford University. 450 Jane Stanford Way, Stanford, CA 94305\\
    $^\mathcal{N}$ Nko USA Inc. 365 E 169th St. Bronx, NY, US 10456\\
    $^\mathcal{F}$ Friasoft. 
9C5M+33, Fria, Guinea.\\
    $^\mathcal{K}$ Kofi Annan University. J986+7P Conakry, Guinea
     }

\if false

\And Baba Mamadi Diané \\ babamamadidiane@gmail.com
\And Solo Farabado Cissé \\ solofarabado@gmail.com
\And Djibrila Diané \\ dianenfakounba@gmail.com
\And Abdoulaye Sow \\ abdoulaye.sow.1989@gmail.com
\And Séré Moussa Doumbouya \\ dseremoussa@gmail.com
\And Daouda Bangoura \\ bangouradaouda@gmail.com
\And Fodé Moriba Bayo \\bayofm@gmail.com
\And Chris Piech \\
\And Christopher Manning \\ manning@cs.stanford.edu

\fi

\begin{document}
\maketitle

% 2009 + 6193 = 8202
\newcommand{\systemname}{Fria$\parallel$el}

\begin{abstract}
Currently, there is no usable machine translation system for Nko \footnote{Also spelled N'Ko, but speakers prefer the name Nko.}, a language spoken by tens of millions of people across multiple West African countries, which holds significant cultural and educational value.
To address this issue, we present a set of tools, resources, and baseline results aimed towards the development of usable machine translation systems for Nko and other languages that do not currently have sufficiently large parallel text corpora available.
(1) \systemname{}: A novel collaborative parallel text curation software that incorporates quality control through copyedit-based workflows.
(2) Expansion of the FLoRes-200 and NLLB-Seed corpora with 2,009 and 6,193 high-quality Nko translations in parallel with 204 and 40 other languages.
(3) nicolingua-0005: A collection of trilingual and bilingual corpora with 130,850  parallel segments and monolingual corpora containing over 3 million Nko words.
(4) Baseline bilingual and multilingual neural machine translation results with the best model scoring 30.83 English-Nko chrF++ on FLoRes-devtest.
\end{abstract}

% input to avoid page break
\section{Introduction}

The Manding languages, including Bambara, Maninka, Mandinka, Dyula, and several others, are generally mutually intelligible and spoken by over 40 million people across West African countries including Mali, Guinea, Ivory Coast, Gambia, Burkina Faso, Sierra Leone, Senegal, Liberia, and Guinea-Bissau. Nko, which means `I say' in all Manding languages, was developed as both the Manding literary standard language and a writing system by Soulemana Kanté in 1949 for the purpose of sustaining the strong oral tradition of Manding languages \citep{niane1974histoire,conde2017, ethnologue2023}.\footnote
  {ISO-639 code:~nqo; ISO-15924 code:~Nkoo.}
Nko thus serves a role for the Manding languages somewhat akin to Modern Standard Arabic for Arabic languages. It adequately transcribes their essential features such as vowel length, nasalization, and tone \cite{oyler2002, conde2017, donaldson2017clear} and enables the development of a shared literature.

% The aim was a standardized language and writing system, which could serve a similar role to Modern Standard Arabic with respect to various regional Arabic languages. Manding languages, which include Mandinka and Bambara, are a subgroup of the Mande language family and are generally mutually intelligible to speakers. Bambara, transcribed in the Latin script, is currently the best-supported Manding language, available in Google Translate and in NLLB-\textsc{Seed}. Our Nko translators are also fluent in Bambara.

Since its invention, the use of Nko has been growing. It is taught by literacy promotion associations, and used in newspapers, social media, and electronic communication \cite{rfi_mandenkan_2016,rosenberg_everyone_2011, Donaldson2019LinguisticAC, kanjamadi_kanjamadi_2022}.
Given that students learn best in their native language \cite{soh_impact_2021}, 
Nko is particularly valuable for elementary native language education.
Unfortunately, Nko and more generally West African languages remain marginalized in West African academic institutions \cite{kotey1975official,bryant2020education}. As a result, and despite the efforts of its courageous community, few academic resources are available in Nko.

Amongst numerous other benefits, computer-assisted translation could be used to facilitate the translation of academic content between Nko and other languages and facilitate projects such as Nko Wikipedia, which currently contains less than two thousand articles, in contrast with French and English Wikipedia with over 2 and 6 million articles respectively \cite{listofwikipedias}. Unfortunately, to date, there isn't any usable machine translation (MT) system for Nko, in part due to the unavailability of large text corpora required by state-of-the-art neural machine translation (NMT) algorithms. 

Nko is a representative case study of the broader issues that interfere with the goal of universal machine translation. Thousands of languages still don't have available or usable MT systems, mainly due to the unavailability of high-quality parallel text corpora.
Recent corpora curation efforts have also resulted in sub-standard data quality for some languages. Some issues reported by \cite{nllb2022} and others that we address in this work (see Section \ref{nllb_seed_alignment}, and \ref{bam_latn_quality}) could have been avoided with the use of an adequate parallel text corpus curation system, which did not previously exist.

This work aims to bootstrap the development of MT systems for Nko and, in the process, to contribute open-sourced resources and tools applicable to other languages. Our main contributions include:

\paragraph{Novel Parallel Text Curation Software.}
Our first contribution is  Fria$\parallel$el (pronounced Friallel), a cloud-based collaborative parallel text curation software that helps human translators orchestrate copy-editing processes resulting in high-quality corpora. Fria$\parallel$el is presented in Section \ref{sec:friallel}.

\paragraph{Extension of FLoRes-200 and NLLB-\textsc{Seed}.}
Our second contribution is the extension of FLoRes-200 and a multilingually aligned version of the NLLB-\textsc{Seed} \cite{nllb2022} corpora with high-quality Nko translations performed by Nko native speaker experts. Both FLoRes-200 and NLLB-\textsc{Seed} match our educational objective fairly well. Both are built over sentences drawn from Wikipedia, with NLLB-\textsc{Seed}, in particular, 
covering various fields of human knowledge and activity. They are therefore more diverse than other common parallel texts, such as religious texts.

\paragraph{Language Resource from the Nko Community.}
Our third contribution is the \emph{nicolingua-0005} corpus, a collection of mono-, bi-, and trilingual corpora curated from data files donated by Baba Mamadi Diané, Solo Farabado Cissé, Djibrila Diané, Nafadji Sory Condé, and Kalo Mory Diané.

\paragraph{Baseline Machine Translation Results.}
Our fourth and last contribution consists of baseline NMT experiments from English, French, and Bambara transcribed in Latin script to Nko and vice versa. We present bilingual and multilingual transformer-based NMT systems \cite{vaswani2017attention} built using the fairseq toolkit \citep{ott-etal-2019-fairseq}. At present, results remain quite modest, with the best \emph{eng\_Latn $\rightarrow$ nqo\_Nkoo} system scoring 30.83 chrF++ on FLoRes-devtest.

All presented software and tools have been publicly released to facilitate further progress on machine translation for Nko and other languages.\footnote
  {Corpora and software on \url{https://github.com/}:\\
    \href{https://github.com/common-parallel-corpora/friallel}{\texttt{common-parallel-corpora/friallel}}\\
    \href{https://github.com/common-parallel-corpora/common-parallel-corpora}{\texttt{common-parallel-corpora/common-parallel-corpora}}\\
    \href{https://github.com/mdoumbouya/nicolingua-0005-nqo-nmt-resources}{\texttt{mdoumbouya/nicolingua-0005-nqo-nmt-resources}}\\
    \href{https://github.com/mdoumbouya/nko-nmt-wmt-2023}{\texttt{mdoumbouya/nko-nmt-wmt-2023}}
}

\section{Fria$\parallel$el: Collaborative Parallel Corpus Curation System}
\label{sec:friallel}

\begin{figure*}
    \centering
    \includegraphics[width=0.95\textwidth]{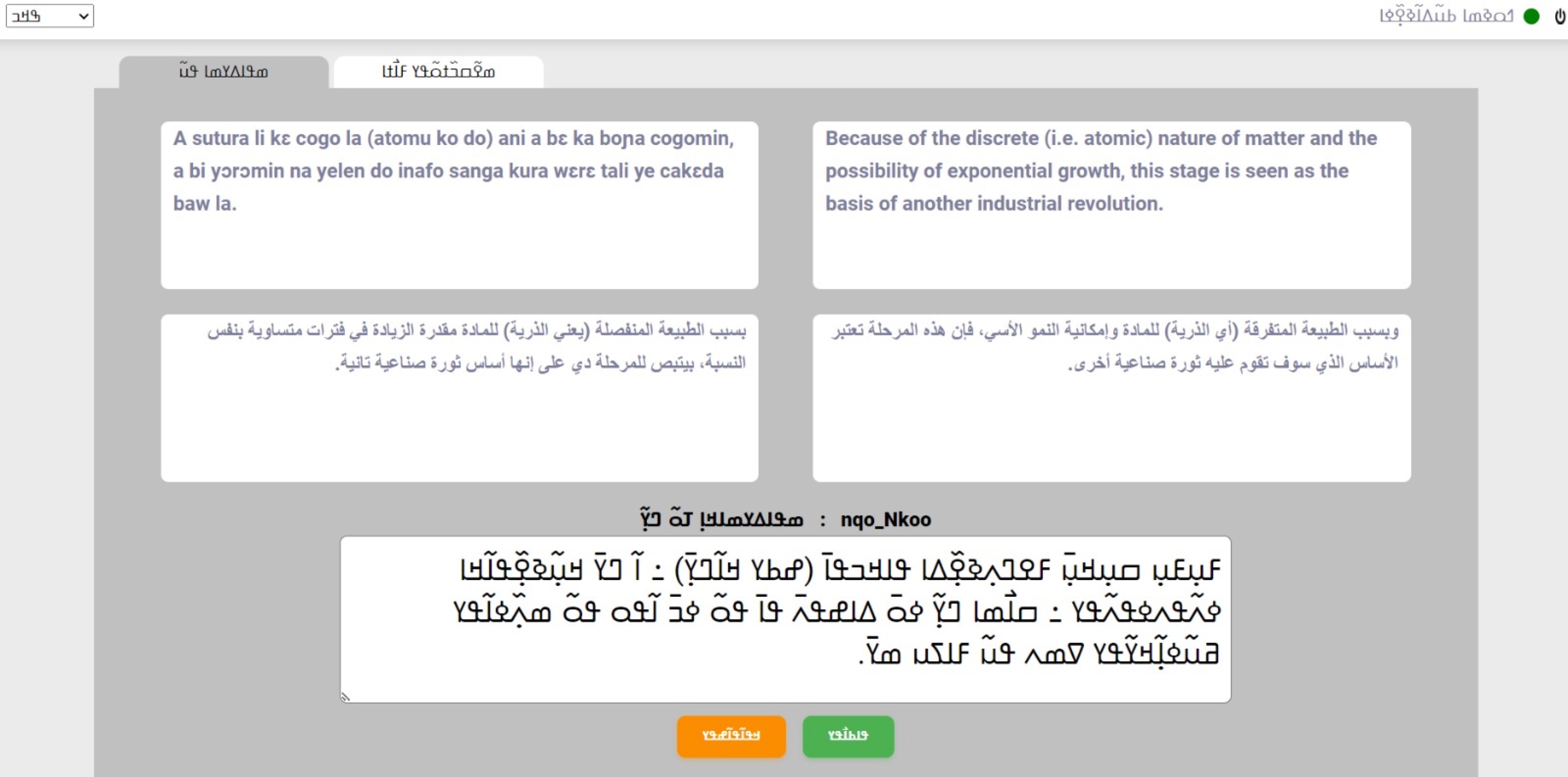}
    \caption{\systemname{}'s user interface for a Nko translator simultaneously inspecting multiple parallel variants of the same segment from the Multitext-NLLB-\textsc{Seed} corpus.
    All labels are localized to Nko. 
    The source language fields are also localized to their own language's writing direction: LTR for Bambara in Latin script and English; and RTL for Moroccan Arabic and Egyptian Arabic. The translated text is localized to Nko's writing direction (RTL).}
    \label{fig:translating-screenshot}
\end{figure*}

% Thousands of languages do not currently have MT technology available, mainly due to the lack of adequate parallel corpora. Available corpora for languages that currently have few resources also tend to be of low quality, further hindering progress.

Recent efforts on collecting multilingual parallel corpora involved sets of data file exchange between various translation teams \cite{federmann2022ntrex}. This process is error-prone as it doesn't allow the systematic tracking of individual corpus entries through a curation quality process. Other recent similar efforts such as NLLB-\textsc{Seed} \cite{nllb2022}, unnecessarily resulted in bi-text data rather than the intended multi-text because the reference files ended up being modified and re-ordered by various translation teams (see Section \ref{nllb_seed_alignment}).
Adequate software could have helped avoid such issues.

We propose \systemname{}, a collaborative system designed to help distributed translation teams produce large multilingually aligned high-quality parallel text corpora. The system design particularly emphasizes suitability for use in various contexts, supporting web and mobile device usage and use in an offline mode. Its design goals include: itemized curation, automatic work organization, collaborative copyediting, and localization to translators' preferred user-interface language and preferred source languages to translate from (Figure \ref{fig:translating-screenshot}).

% NOTES
% (1) target audience
% (2) bootsrapping MT for languages with few resources
% (3) easily extensible workflows
% (4) translator skill levels (l1, l2, l3 for nqo)
% (5) extensible platform: e.g. syntax annotation

\subsection{Previous Tools and Multilingual Parallel Corpora Creation Processes}
\paragraph{Masakhane}
Similarly to \cite{nekoto-etal-2020-participatory}, this work is an effort towards African language technology development. Our work is participatory in the sense that we are a diverse team of computer scientists, linguists, and native speakers of Nko and other West African languages. We expect that our approach, and the parallel text curation software we release with this paper, \systemname{}, will be valuable for MT technology development for other languages.

\paragraph{ParaText} ParaText \cite{paratext} is specialized software for Bible translation projects. Its features include team management, task assignments, notes, collaborative document editing, multilingual dictionaries, and various biblical resources. It also allows a side-by-side comparison of biblical passages from various sources or in various languages. 
Paratext is not suited for general-purpose parallel corpus curation for MT. There is no indication that ParaText or any such software was used in the curation process of recent multilingual parallel corpora such as NLLB-\textsc{Seed}, FLoRes-200, and NTREX-128.

\paragraph{NLLB-\textsc{Seed} and FLoRes-200}
The curation process of FLoRes-200 involved teams of translators and reviewers who underwent a vetting process. The QA team reviewed a 20\% subset of data files with  3000 entries produced by translation teams. Data files falling below the 90\% quality threshold were returned for rework. NLLB-\textsc{Seed} underwent a less rigorous quality control process.
The curation process was English-centric. Translators were required to be proficient in English. Translation to the majority of languages was also done from English, with the following exceptions: In NLLB-\textsc{Seed}, Ligurian, was translated from Italian, In FLoRes-200, some Arabic languages were translated from Modern Standard Arabic. As noted by the authors, there are qualified translators who may not speak English, and several languages may be easier to translate from non-English sources.

\paragraph{NTREX-128}
NTREX-128 \cite{federmann2022ntrex} was curated as follows. The English reference file was sent to a translation provider that produced translations. Source-based direct assessment was performed on the translated files by a different provider using the Appraise platform \cite{federmann-2018-appraise} to generate segment-level quality scores. Segments with a score below a specified threshold were returned for correction. The translation process and quality control method of the translation provider were not specified.

\systemname{}  is a collaborative parallel text curation software system that tracks individual segments through a translation and copyedit workflow. Each segment is translated by one translator, and subsequently sequentially copyedited by other translators. \systemname{} allows translators to simultaneously inspect variants of the source segment in multiple languages. This results in segments translated and copyedited in the context of different subsets of source languages.
In addition to the final parallel corpus, \systemname{} also yields copyedit logs, which could be valuable in various modeling scenarios.

\subsection{Design Goals}
\label{sec:design-goals}
\systemname{} was designed with the following goals:

\paragraph{Itemized Curation}
Each corpus segment is individually tracked through the curation process in which it is translated to the target language and subsequently reviewed and copyedited several times.

\paragraph{Automatic Task Assignments}
Translation and copyediting tasks are automatically assigned to translators with fixed lease periods. Uncompleted tasks are automatically reassigned upon expiration.

\begin{figure}
    \centering
    \includegraphics[width=0.45\textwidth]{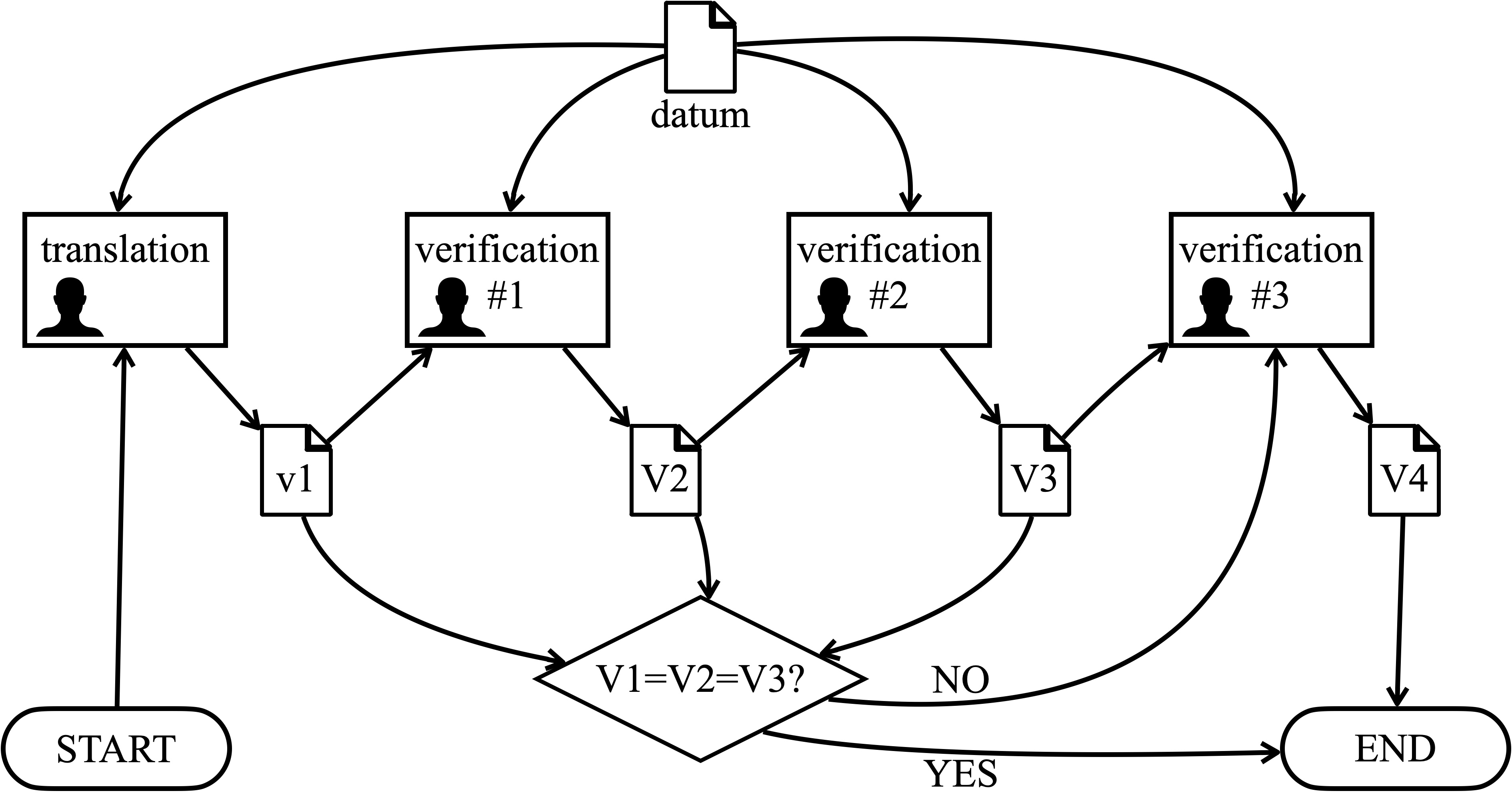}
    \caption{Translation workflow for a multilingual segment (datum). The initial translation (\emph{v1}) is approved or copyedited by two other translators (\emph{v2}) and (\emph{v3}). If any copyediting occurs, a third copyediting task is assigned to a fourth translator who either approves the current translation or performs a final copyedit (\emph{v4}).}
    \label{fig:data-curation-workflow}
\end{figure}

\paragraph{Collaborative Copyediting}
Each segment is translated once and copyedited two or three times, following the workflow in Figure~\ref{fig:data-curation-workflow}. Segments for which the first or second verification results in edits are copyedited a third time. A given translator can only perform a task on a given segment once.

\paragraph{Multilingual Sources}
While performing translation and copyediting tasks, translators can simultaneously inspect segments in several languages configured according to their preferences.

\paragraph{Machine-Generated Sources}
Datasets can be augmented with additional machine-generated variants of segments such as machine translations, transliterations, and detransliterations.

\paragraph{Responsive Web Design}
\systemname{} is a web application that automatically adapts the layout of its component to the user's screen size. This makes it usable on desktop and laptop computers as well as on mobile phones and tablets.

\paragraph{Resilience to Connectivity Disruptions}
Translators who temporarily lose their internet connectivity can seamlessly keep working offline on their currently assigned translation and verification tasks. Their work is automatically synchronized with the central database when their connectivity is restored.

\paragraph{Internationalization and Localization} 
\systemname{} is internationalized (i18n) in that all user-facing strings are externalized into a translatable resource file, and
the writing direction and text alignment of translation source and target languages are configurable.
As a result, the user interface is localized (L10n) to the translator's preferred user-interface language, and to each source language (Figure \ref{fig:translating-screenshot}).

\subsection{Software Components}
This section provides details on \systemname{}'s software components that collectively realize the design goals specified in Section \ref{sec:design-goals}.

% (ISO 639: nqo, ISO 15924: Nkoo)

\subsubsection{Workflow Manager}
\label{sec:workflow-manager}
Both the Workflow and Task Managers are implemented as Firebase cloud functions that are triggered at fixed time intervals. A workflow entity is inserted for each parallel segment with an initial \emph{active} state.
The Workflow Manager periodically inspects workflow entities and (1) creates the next task if needed, and per the workflow management rules, (2) moves the workflow to the \emph{completed} status if all related tasks have been completed and there is no need to create additional tasks or (3) nothing, if the workflow has an uncompleted task.

\subsubsection{Task Manager}
\label{sec:task-manager}
When triggered, the Task Manager revokes all expired task assignments and assigns unassigned translation and copyedit tasks to users according to their roles. The maximum number of tasks assigned to each user is fixed.
A given user is never assigned a task related to a segment on which they have previously completed a task.
The Task Manager also ensures that a copyedit task is only assigned to a user with the appropriate verification skill level (\emph{L1}, \emph{L2}, or \emph{L3}) for the first, second, and third copyedit rounds. 
Each translator account is configured with specific verification skill levels.

\subsubsection{Data Model and Storage}
Google Firestore, a document-oriented NoSQL database, is used for data storage. The central application database is accessed by data import/export scripts, the WorkflowManager, the Task Manager, and the user interface. It contains the following collections of documents:

\textbf{datasets:} One collection per imported dataset. Each document represents a multilingual segment and contains all available translations of the segment, each annotated with its language and writing system. See Figure~\ref{fig:corpora-storage}.

\textbf{workflows:} Each document represents a prioritized workflow entity. The WorkflowManager (Section~\ref{sec:workflow-manager}) periodically inspects workflow entities by priority order and creates task entities as per the workflow management logic.

\textbf{annotation-tasks:} Each document is a task of a specific type (translation or copyedit) related to a specific multilingual segment. Each task has a status (unassigned, assigned, completed). Tasks are assigned to translators by the Task Manager.

\textbf{users:} Each document represents a translator and specifies whether they can be assigned translation (\emph{isActiveTranslator}) and copyedit (\emph{isActiveVerifier}) tasks. Translator documents also store the source languages the translator prefers to translate from, subject to availability in the source  corpus. User documents also specify a \emph{verifierLevel}, which indicates the maximum copyediting round the translator can participate in for a specific segment.

\textbf{config:} Contains language writing direction configuration. Languages are assumed to be left-to-right unless explicitly marked right-to-left.

\subsubsection{User Interface}
% TODO BAYO
The user interface is a responsive web application that is usable on a variety of devices, including mobile phones, tablets, desktops, and laptops (Figure \ref{fig:translating-screenshot}). It directs authenticated translators to their workspace where they can perform translation (first tab) and copyediting (second tab) tasks that are assigned to them. The task assignment process is transparent to translators. One task is displayed at a time. The prioritized list of tasks assigned to the current translator is kept in a cache for resilience to intermittent internet disruptions. The connection status is indicated by the green circle (top-right).

When performing translation tasks, translators simultaneously inspect the source segment in several languages (top four text fields) and write a translation in the target language text field (bottom). When the `submit' button (green) is selected, the translation is recorded and the next task is displayed. Translators can also skip the current task by selecting the `skip' button (orange). When performing copyediting tasks, the bottom text field is initialized with the latest version of the translated segment (Figure \ref{fig:data-curation-workflow}). The translator may leave the translation intact or copyedit it before submitting.

\subsubsection{Offline Mode}
\begin{figure}
    \centering
    \includegraphics[width=0.45\textwidth]{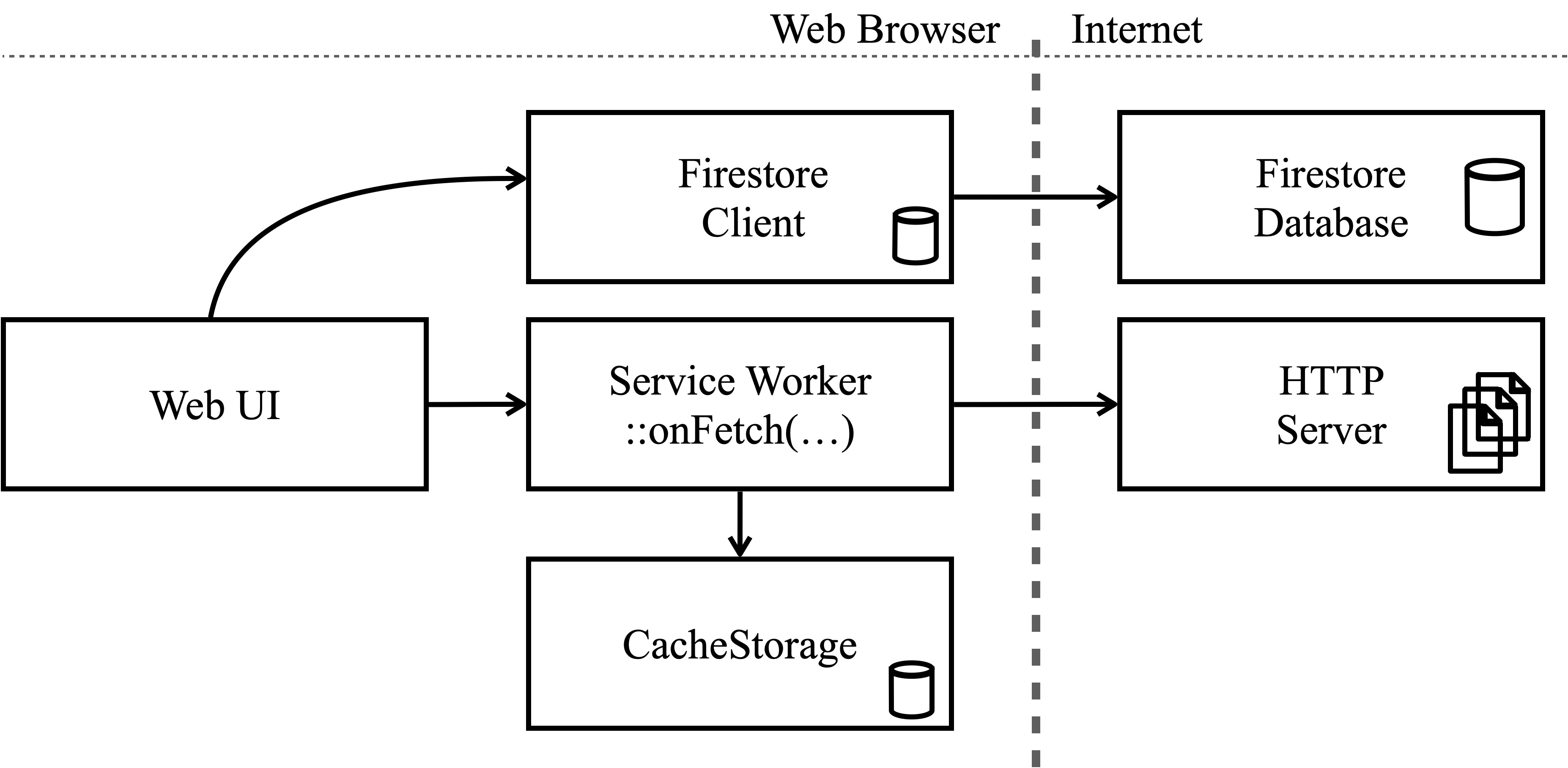}
    \caption{The software uses Firestore's client library's offline mechanism and  \emph{cached-aside} HTTP resources to be resilient to intermittent internet disruptions.}
    \label{fig:offline-architecture}
\end{figure}
The software is a web application designed to be resilient to intermittent internet disruptions. This is achieved with Google Firebase's client library \cite{firebase_offline}, which supports offline read and write operations by leveraging a client-side \emph{eventually consistent} \cite{burckhardt2014principles} LRU cache, and \emph{cached-aside} \cite{pamula2014cache} HTTP resources, implemented with
two web APIs supported by the majority of web browsers: CacheStorage and ServiceWorker \cite{w3_service_workers, web_api_service_worker, web_api_cache_storage}.
After the initial loading of the web application in a web browser, a ServiceWorker is registered to intercept HTTP fetch events. If the remote web server is reachable, the ServiceWorker fetches remote HTTP resources (e.g., HTML, CSS, javascript, image files) and stores them in a CacheStorage before returning them to the caller; otherwise, cached resources are served. The entire process is transparent to the user. See Fig~\ref{fig:offline-architecture}.

\subsubsection{Translator Copyedit Logs}
In addition to the final version of the translated segments, the data \systemname{} also outputs their initial translation (\emph{v1}), and the versions of the same entries after the first, second, and third copyediting rounds (\emph{v2}, \emph{v3}, \emph{v4}) -- see the workflow in Figure~\ref{fig:data-curation-workflow}. Copyediting logs can be valuable in developing language and machine translation models.

\subsubsection{Data Import and Export}
\systemname{} includes the following administrative Python scripts for importing and exporting parallel corpora and other reports.
\verb|load_dataset.py| imports a new parallel corpus from its original data files. Pre-processing may be required to adapt to various original dataset formats.
\verb|create_translation_workflows.py| creates active translation workflows for an imported dataset.
\verb|system_report.py| displays the number of workflows and tasks by status by dataset.
\verb|export_dataset.py| exports translations and translator edits for a curated dataset in a csv file. Post-processing may be required to adapt to a desired format.
\verb|accounting_statements.py| generate completed tasks by user by dataset by month. This data can be imported into an accounting system to generate payroll for translators.

\subsection{Qualitative User Study}
Nko translators used \systemname{} to translate  FLoRes-200 (dev, devtest) and Multitext-NLLB-\textsc{Seed} to nqo\_Nkoo, and to copyedit each segment two or three times.
The following sections present an analysis of their responses to a survey questionnaire (Figure \ref{fig:survwy-questionnaire}).
Quantitative measures on their copyediting logs are also discussed in Section \ref{copy-editing-metrics}.

\subsubsection{Usability}

Nko translators praised the simplicity of the user interface. They appreciated the automatically organized itemized copyediting-based data curation process.
They highlighted the localization features, particularly, the fact that the user-interface is available in Nko and that the presentation was adequate for both right-to-left and left-to-right source languages and the target language.
They valued the offline functionality that allowed them to temporarily continue working without an internet connection.
Furthermore, they found the task counters displayed on the user-interface helpful.
They noted two usability-related limitations: First, it was not possible to directly go back to a task after submitting it. Second, although the software allowed them to continue working offline, it did not allow them to perform the initial authentication while offline.

\subsubsection{Translation Process}

Nko translators found the fact that source segments were visible in multiple languages beneficial. They said that the ability to inspect the same segment in multiple languages facilitated its translation to Nko.
They also mentioned that the itemized translation tasks, which presented one segment at a time, decreased the likelihood of translation mistakes.

An improvement they requested is the addition of a translation memory including dictionary entries and previously translated expressions.

\subsubsection{Copyediting Process}
Nko translators found \systemname{}'s multi-pass copyediting process effective for finding and correcting translation mistakes.
They mentioned that the fact that segments were consecutively assigned to different translators for copyediting led to higher-quality translations as it is easy to overlook one's own mistakes.
Because each translator had a different translation source language configuration, 
Nko segments were translated from and copyedited against their versions in different sets of languages, which Nko translators found enriching.

\subsubsection{Mistranslations}
Types of mistranslations Nko translators noted during the copyediting process included typos, omitted words, grammatical errors, incorrect word sense translations, incorrect translations of named entities, and punctuation errors. They noted that word sense was sometimes hard to disambiguate without the full context of segments. For instance, the English word \emph{state} maps to different Nko words based on the sense of the word (political community vs.\ a particular condition of a person, place, or thing).
They also noted punctuation errors, particularly the use of the Arabic comma (U+060C) instead of the Nko comma (U+07F8), and spacing around that punctuation.
Finally, they reported that translators using different source languages would sometimes disagree on named entity translations.

\subsubsection{Disagreements}
Nko translators reported few disagreements on language standards. They also reported using existing English-Nko and French-Nko dictionaries for consistency.
During the translation of FLoRes-200, NLLB-\textsc{Seed} to Nko, translators participated in weekly team meetings and routinely consulted each other over video conferences and phone calls. They deferred the few cases of disagreement and perplexing questions to the most senior translator.

%\paragraph{Questions}
%\begin{itemize}
%    \item How were disagreements on foreign-%named entity translations resolved?
%    \item How were words missing in Nko %dictionaries handled?
%\end{itemize}

\subsubsection{Copyediting Metrics}
\label{copy-editing-metrics}
Table~\ref{tab:edit-statistics} summarizes the size of the translated corpora in segments and Nko words, as well as the percentage of segments that were edited in each verification round, and the related edit magnitudes, computed as edit distances.
The number of edited segments and related edit magnitudes generally decreased as copy-editing rounds progressed.

\begin{table*}[ht!]
    \centering
    \begin{tabular}{@{}lrr|rr|rr|rr@{}}
    \toprule
        ~ & \multicolumn{1}{l}{seg-} & ~ & \multicolumn{2}{c|}{$v1 \rightarrow v2$} & \multicolumn{2}{c|}{$v2 \rightarrow v3$} & \multicolumn{2}{c}{$v3 \rightarrow v4$} \\ 
        corpus & ments & words &edited & edit distance & edited & edit distance & edited & edit distance \\ \midrule
        FLoRes-dev& 997 & 27,361 & 83\% & $38.75 \pm 1.55$ & 67\% & $50.48 \pm 2.10$ & 71\% & $11.74 \pm 0.65$ \\
        % \hline
        FLoRes-devtest& 1,012 & 29,503 & 87\% & $61.74 \pm 1.81$ & 93\% & $9.69 \pm 0.64$ & 24\% & $2.79 \pm 0.15$\\
        % \hline
        NLLB-\textsc{Seed}& 6,193 & 184,138 & 48\% & $45.97 \pm 1.11$ & 35\% & $38.94 \pm 1.16$ & 35\% & $11.96 \pm 0.48$\\
    \bottomrule
    \end{tabular}
    \caption{Percentage of edited Nko segments, and related mean$\pm$ standard error of edit magnitudes (edit distance) resulting from the translation of FLoRes-dev, FLoRes-devtest, NLLB-\textsc{Seed} to Nko}
    \label{tab:edit-statistics}
\end{table*}

\section{Nko Corpora for Machine Translation}
\label{sec:parallel-corpora}
This section discusses the extension of FLoRes-200 and NLLB-\textsc{Seed} to Nko, which included the multilingual alignment of NLLB-\textsc{Seed}, and the use of \systemname{} to translate those corpora to Nko.
This section also introduces \emph{nicolingua-0005}, a collection of monolingual corpora and bi- and trilingual parallel corpora donated by Nko community members.

\subsection{Translation of FLoRes-200 and NLLB-Seed to Nko}
Nko native speaker experts Baba Mamadi Diané, Solo Farabado Cissé, and Djibrila Diané, used \systemname{} to translate Multitext-NLLB-\textsc{Seed}, FLoRes-dev, and FLoRes-devtest to Nko.
They worked from Cairo (Egypt), Banankoro (Guinea), and New York (USA), and with the rest of the team, participated in weekly video conference meetings.

\subsection{Translation Process}
The initial translations of FLoRes-dev, which our translators performed using spreadsheets, were imported into \systemname{} after our software engineers completed its development. The copyediting tasks for FLoRes-dev, and  the translation and copyediting tasks for  FLoRes-devtest and Multitext-NLLB-\textsc{Seed} were entirely performed using \systemname{}. The system was designed not to allow translators to copyedit their own translations or previous copyedits. This constraint made the proposed translation workflow impossible given the size of our team of translators. As a workaround, an additional user account was created for the two most experienced translators to allow third copyediting rounds.

Each segment was translated once and copyedited two or three times. The resulting curated Nko data files are summarized in Table~\ref{tab::dataset}. The multilingually aligned NLLB-\textsc{Seed} dataset (Multitext-NLLB-\textsc{Seed}), FLoRes-dev, and FLoRes-devtest, all extended with Nko translations along with copyedit logs, collectively make up common-parallel-corpora ver. 2023-06-19 summarized in Table~\ref{tab:common-parallel-corpora}.

\begin{table}
    \centering
    \begin{tabular}{rrl}
        \toprule
        \multicolumn{3}{c}{Translations} \\
        \midrule
        lines & words & file \\ 
        \midrule
        6193 & 184138 & \textsc{Seed}/nqo\_Nkoo \\ 
        997 & 27361 & FLoRes/nqo\_Nkoo.dev \\
        1012 & 29503 & FLoRes/nqo\_Nkoo.devtest \\
        \bottomrule
        \\
        \toprule
        \multicolumn{3}{c}{Translator Edits} \\
        \midrule
        lines & words & file \\ 
        \midrule
        6193 & 170555 & \textsc{Seed}/nqo\_Nkoo.v1 \\ 
        6193 & 177703 & \textsc{Seed}/nqo\_Nkoo.v2 \\ 
        6193 & 182843 & \textsc{Seed}/nqo\_Nkoo.v3 \\ 
        6193 & 184138 & \textsc{Seed}/nqo\_Nkoo.v4 \\ 
        \hline
        997 & 24455 & FLoRes/nqo\_Nkoo.dev.v1 \\ 
        997 & 25656 & FLoRes/nqo\_Nkoo.dev.v2 \\ 
        997 & 26541 & FLoRes/nqo\_Nkoo.dev.v3 \\ 
        997 & 27361 & FLoRes/nqo\_Nkoo.dev.v4 \\ 
        \hline
        1012 & 25924 & FLoRes/nqo\_Nkoo.devtest.v1 \\ 
        1012 & 27771 & FLoRes/nqo\_Nkoo.devtest.v2 \\ 
        1012 & 29521 & FLoRes/nqo\_Nkoo.devtest.v3 \\ 
        1012 & 29503 & FLoRes/nqo\_Nkoo.devtest.v4 \\ 
        \bottomrule
    \end{tabular}
    \caption{Extensions of FLoRes-200 (dev, devtest) and Multitext-NLLB-\textsc{Seed} to Nko. The \emph{nqo\_Nkoo} data files are parallel with 40 other languages in NLLB-\textsc{Seed}, and 204 other languages in FLoRes-200. FLoRes-test, which is not publicly available, was not translated.}
    \label{tab::dataset}
\end{table}

\begin{table}[]
    \centering
    \begin{tabular}{lrrr}
    \toprule
    CPC subset & lines & langs. & tr. edits \\
            &           &       & langs. \\
    \midrule
    Multitext-NLLB-\textsc{Seed} & 6193 & 41 & 1 \\
    FLoRes-dev & 997 & 205 & 1 \\
    FLoRes-devtest & 1012 & 205 & 1 \\
    \bottomrule
    \end{tabular}
    \caption{Summary of common-parallel-corpora version 2023-06-19. All entries are parallel across all languages. Translator edits are only available for nqo\_Nkoo.}
    \label{tab:common-parallel-corpora}
\end{table}

\subsection{Multilingual Alignment of NLLB-\textsc{Seed}}
\label{nllb_seed_alignment}
% The original NLLB-\textsc{Seed} dataset has a limitation: entries in non-English languages are only parallel with English.
% Furthermore, the English corpus differs between the included bilingual subsets by minor edits, and by line orderings.
The original NLLB-\textsc{Seed} dataset consists of pairwise parallel corpora between English and each other language but suffers from the complication that many of the source English sides are slightly different from each other, variously due to minor copyediting, and reordered and added entries.

Multitext-NLLB-\textsc{Seed} is a multilingually aligned version of NLLB-\textsc{Seed} that fixes this limitation. It was created as follows:
A consensus \emph{eng\_Latn} reference file was manually edited by human comparison of all existing reference \emph{eng\_Latn} files.
The lines of each \emph{eng\_Latn} file were matched (binary assignment matrix $M_{i,j}$) to the lines of the consensus \emph{eng\_Latn} file by minimizing the sum of the edit distances $E_{i,j}$ between matched lines (Equation \ref{eq:min-sum-assignment}). The optimal line matching $M^*$, obtained using the scipy package \cite{2020SciPy-NMeth}, was used to re-order each non-English language file to match the order of the consensus \emph{eng\_Latn} file.
Two unmatched lines from (\emph{eng\_Latn},~\emph{kas\_Deva}) and one from (\emph{eng\_Latn},~\emph{lij\_Latn}) were discarded.

%
%6195 eng\_Latn-kas\_Deva
%6194 eng\_Latn-lij\_Latn
%6193 (all other files)
%   
\begin{equation}
    M^* = \arg\min_{M} \sum_{i,j}M_{i,j}E_{i,j}
    \label{eq:min-sum-assignment}
\end{equation}

\begin{figure}
    \centering
    \includegraphics[width=0.5\textwidth]{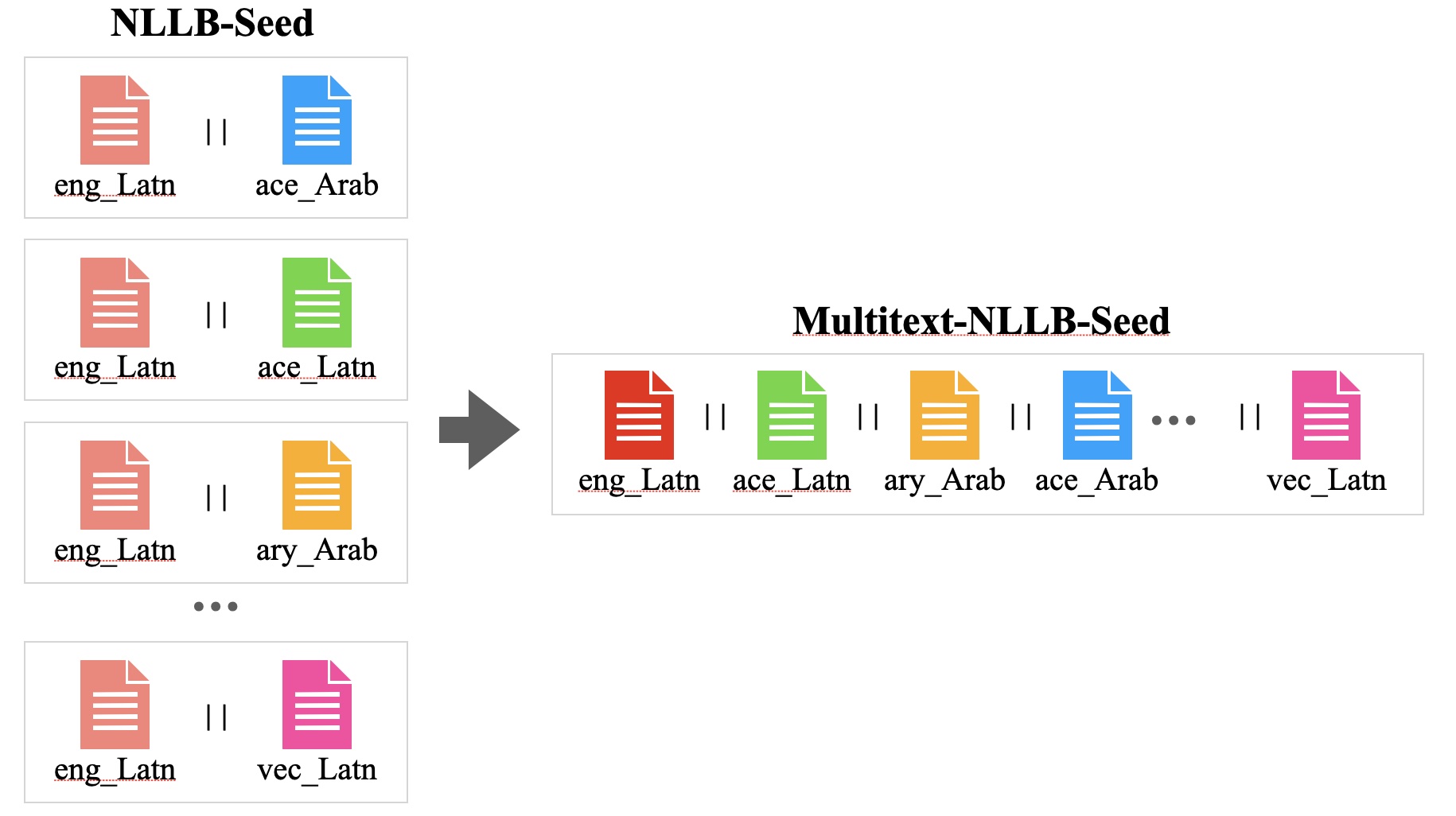}
    \caption{Multitext-NLLB-\textsc{Seed} is a multilingually aligned version of the original NLLB-\textsc{Seed} dataset.}
    \label{fig:multitext-nllb}
\end{figure}

The resulting re-ordered non-English language files and the consensus eng\_Latn file constitute the Multitext-NLLB-\textsc{Seed} corpus, containing 40 parallel language files; see Figure~\ref{fig:multitext-nllb}. Multitext-NLLB-\textsc{Seed} was loaded in \systemname{} in lieu of the original NLLB-\textsc{Seed} corpus, enabling translators to inspect each segment in multiple languages, and resulting in an expanded multilingually aligned corpus.

\subsection{Translation Source Languages}
Source languages were configured in \systemname{} according to the preferences of each translator. Collectively, they translated 
 % FLoRes-200 and Multitext-NLLB-\textsc{Seed} 
 from
\emph{fra\_Latn}, 
\emph{eng\_Latn}, 
\emph{ary\_Arab}, 
\emph{arz\_Arab}, and 
\emph{bam\_Latn}. Note that \emph{fra\_Latn} is not available in NLLB-\textsc{Seed}. \emph{bam\_Nkoo} was detransliterated from \emph{bam\_Latn} using a neural detransliterator \cite{Doumbouya_Detransliterator_2022}; however, translators did not find this source useful and preferred not to enable it in their configuration.

\subsection{Further Notes on Manding languages}
Nko was developed as a standardized form of the Manding languages. The aim was a standardized language and writing system, which could serve a similar role to Modern Standard Arabic with respect to various regional Arabic languages. Manding languages, which include Mandinka and Bambara, are a subgroup of the Mande language family and are generally mutually intelligible to speakers. Bambara, written in a Latin script, is currently the best-supported Manding language, available in Google Translate and in NLLB-\textsc{Seed}. Our Nko translators are also fluent in Bambara.

\subsection{Quality of bam\_Latn in NLLB-\textsc{Seed}}
\label{bam_latn_quality}
Our 
Nko translators noted the following quality issues with NLLB-\textsc{Seed}'s \emph{bam\_Latn} data:
(1) The data contains too much French vocabulary not enough Manding vocabulary. 
(2) Some entries do not match their English counterpart at all. 
(3) Some entries are entirely in French; examples are shown in Figure~\ref{fig:bam_latn_quality_issues}.
(4) The \emph{bam\_Latn} data completely lacks tonal marks, which are important in Manding languages (e.g., many nouns are indistinguishable without tonal marks, such as \emph{bird}, \emph{belly}, \emph{inside}; the definite and indefinite inflections of nouns cannot be distinguished without tonal marks (\emph{I saw a person} vs.\ \emph{I did not see any person}); and nouns that can be used as a verb and their verb form cannot be distinguished (\emph{get out!}\ vs.\ \emph{to get out}). \emph{bam\_Nkoo}, detransliterated from \emph{bam\_Latn} was included in the corpus; however, some Nko translators did not find it useful and preferred to not enable it as a source.

% Note on Tonal Marks
% example
% #3161 in NLLB-\textsc{Seed}
% Badjibaw ka érézon fitiniw bé wélé kôgôdjiw, golfouw, djidawolow ani tôgô wèrèw.
% 
% #correction by baba M diane
% Bájíbáù ká érézɔ́n` fítíninù bɛ́ wélé kɔ̀gɔ̀jíú`, kɔ́lfúc, jídáwólóú` à ní tɔ́kɔ́ wɛ́rɛ́ú

\iffalse
Notes on variations of API tonal marks
\begin{verbatim}
ߊ߰ߘߎ߰ ߥߟߊ߫ ߜߊ߰ߣߊ߫ ߞߘߐ߬ߡߊ߲ ߘߐ߬ߝߐ ߓߊߕߐߡߐ߲
Harvard API: Wà:dù: wálá Gà:ná kɔ̀dɔ̀mã dɔ̀fɔ batɔmɔ̃
Library of congress API: wàadùu wálá gbàaná kɔ̀dɔ̀mán` dɔ̀fɔ́` bátɔ́mɔ́n`
\end{verbatim}
\fi

\begin{figure}
    \centering
    \includegraphics[width=0.5\textwidth]{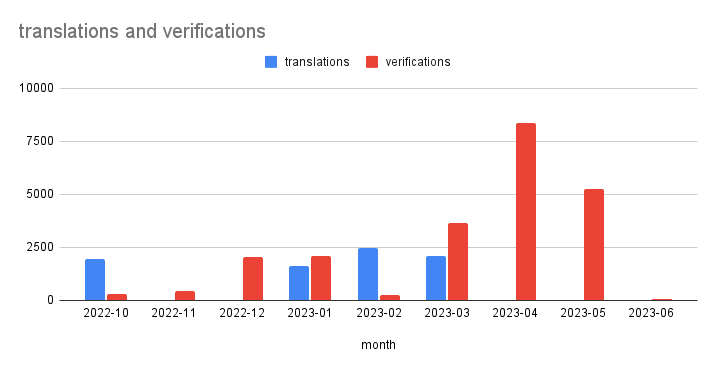}
    \caption{From October 2022 to June 2023, 8,202 translations and 22,426 verifications/edits were performed to produce high-quality translations of FLoRes-200 and Multitext-NLLB-\textsc{Seed} to Nko.}
    \label{fig:verifications}
\end{figure}

% \subsection{TODO}
% \begin{itemize}
%     \item examples of translator edits
%     \item  mean/std edit distance from v1 to v2 to v3 to v4
% \end{itemize}

\subsection{\emph{nicolingua-0005} Corpus}

\emph{nicolingua-0005}
 is curated from files donated by Nko community members for the purpose of developing machine translation for Nko.
It is comprised of 3.9 million Nko words with 25K (Nko, English, French) parallel segments, 59K (Nko, English) parallel segments, 45K (Nko, French) parallel segments, and a monolingual corpus of 3.3 million Nko words.
Included datasets were curated from files provided by Baba Mamadi Diané, Solo Farabado Cissé, Djibrila Diané, Nafadji Sory Condé, and Kalo Mory Diané.
See Table \ref{tab:summary-of-nicolingua-0005}
and Appendix~\ref{sec::nicolingua-0005-details} for more details on the constitution of the corpus. A datasheet questionnaire based on \cite{costa2020mt} is presented in 
Appendix~\ref{sec::datasheet-questionnaire}.

% (ISO 639: nqo, ISO 15924: Nkoo)

\begin{table}
    \centering
    \begin{adjustbox}{max width=0.45\textwidth}
    \begin{tabular}{@{}llrr@{}}
    \toprule
        type & languages & segments & nqo words\\
    \midrule
        trilingual & \emph{nqo\_Nkoo}, \emph{eng\_Latn}, \emph{fra\_Latn} & 25 848 & 256 934\\
        bilingual & \emph{nqo\_Nkoo}, \emph{eng\_Latn} & 59 442 & 283 279\\
        bilingual & \emph{nqo\_Nkoo}, \emph{fra\_Latn} & 45 560 & 129 789\\
        monolingual & \emph{nqo\_Nkoo} & N/A & 3 291 371\\
        \hline
        total & ~ & 130 850 & 3 961 373\\
    \bottomrule
    \end{tabular}
    \end{adjustbox}
    \caption{Summary of \emph{nicolingua-0005}}
    \label{tab:summary-of-nicolingua-0005}
\end{table}

\section{Baseline Machine Translation Experiments}
\label{sec:baseline-nmt-experiments}

This section describes Transformer \cite{vaswani2017attention} based encoder-decoder neural machine translation models built using the fairseq toolkit \citep{ott-etal-2019-fairseq}.
Both bilingual and multilingual translation models are explored. At present, results remain quite modest, with the best model achieving a $30.83$ \emph{eng\_Latn $\rightarrow$ nqo\_Nkoo} chrF++ score on the CPC/FLoRes-devtest corpus.

Eight models were trained: 
The bilingual unidirectional models 200.11 and  200.16, the multilingual model 201.16, and its variant that is trained to also autoencode Nko segments 202.16, and Models 206.19, 207.19, 208.19 and 209.19, which explore three different ways of specifying language tokens.

\subsection{Datasets}
common-parallel-corpora (CPC) and \emph{nicolingua-0005}, described in Section \ref{sec:parallel-corpora} are used to build baseline NMT models for the following translation directions:
\emph{nqo\_Nkoo} $\rightleftarrows$ \emph{eng\_Latn},
\emph{nqo\_Nkoo} $\rightleftarrows$ \emph{fra\_Latn}, and
\emph{nqo\_Nkoo} $\rightleftarrows$ \emph{bam\_Latn}.
The subsets of those corpora used to train, validate, and test the models are specified in Tables \ref{tab:data-details-train} and \ref{tab:data-details-valid-test}.

\subsection{Tokenization}
Byte-pair encoding (BPE) \cite{sennrich-etal-2016-neural} is employed to perform sub-word tokenization. In each training experiment, the BPE model is trained on a token corpus constructed by concatenating all data files containing the languages of interest in the training set. In all cases, the BPE model is trained to produce 15K sub-word units.

\subsection{Models}
Eight models were trained. The first two, 200.11 and 200.16, are unidirectional bilingual \emph{nqo\_Nkoo} $\rightleftarrows$ \emph{eng\_Latn} models. The last six, 201.16, 202.16, 206.19, 207.19, 208.19, and 208.19 are multilingual 
\emph{nqo\_Nkoo} $\rightleftarrows$ \emph{eng\_Latn},
\emph{nqo\_Nkoo} $\rightleftarrows$ \emph{fra\_Latn}, and
\emph{nqo\_Nkoo} $\rightleftarrows$ \emph{bam\_Latn} models.

\subsubsection{Bilingual Models}
200.11 is the baseline bilingual \emph{nqo $\leftarrow$ eng} model. 200.16 differs from 200.11 in terms of model architecture and hyper-parameters. 200.16 and the multilingual models 201.16 and 202.16 have identical architectures and training hyper-parameters.

\paragraph{Model 200.11} is a Transformer-based \cite{vaswani2017attention} encoder-decoder sequence-to-sequence model consisting of 5 encoder and 5 decoder layers, each with a 512-dimensional token embeddings and 2048-dimensional feed-forward networks, 2 attention heads per layer, and a layer normalization module before each layer.
Its architecture and training hyper-parameters are identical to the baseline system of the AmericasNLP 2021 Shared Task on Open Machine Translation \cite{mager2021findings}, except for the following differences: 
(1) encoder and decoder embeddings are not shared,
(2) Subword Regularization \cite{kudo-2018-subword-regularization} and BPE-dropout \cite{provilkov-etal-2020-bpe} are not employed in BPE tokenizer training,
(3) larger batches are employed during training,
(4) gradient clipping is applied during training.

\paragraph{Model 200.16}
This model is only different from 200.11 in that it is deeper (6 encoder layers and 6 decoder layers), and that it is trained with a higher token dropout probability (0.6 instead of 0.4).

\subsubsection{Multilingual Models}
Our multilingual models are trained on parallel corpora obtained by concatenating all available (\emph{nqo} $\rightleftarrows$ \emph{eng}, \emph{nqo} $\rightleftarrows$ \emph{fra}, \emph{nqo} $\rightleftarrows$ \emph{bam}) bitext and prefixing the source segments with language tokens as introduced by \cite{johnson2017google}.
Similarly to \cite{wicks-duh-2022-effects}, models 206.19, 207.19, 208.19, and 209.19 compare the effect of various approaches to constructing source-side prefixes.

\paragraph{Model 201.16} 
is the baseline multilingual model. It has the same architecture and training hyperparameters as the bilingual model 200.16, but it is trained on multilingual data and it employs target language token prefixes (Table \ref{tab:model-lang-tokens}).

\paragraph{Model 202.16}
 employs target language tokens just like 201.16, but its training set also contains  \emph{nqo} $\rightarrow$ \emph{nqo} pairs where each side is the same sentence from monolingual Nko corpora in \emph{nicolingua-0005}. Consequently, 202.16 performs simultaneous multilingual translation and monolingual sequence autoencoding. Positive results from such a strategy were found in \citep{luong2016multi}.

\paragraph{Models 20x.19} also perform simultaneous translation and monolingual sequence auto-encoding. However, their architecture is different from 202.16, and they explore different language token prefixing strategies.
Compared to 202.16 models, in 20x.19 models, the encoder and decoder layers use 8 attention heads instead of 2. Also, the encoder's input embeddings and the decoder's input and output embeddings are all shared. Finally, the source and target token dictionaries are also shared. 

Models 20x.19 explore four approaches of source-side prefix specification (Table~\ref{tab:model-lang-tokens}). As an example, a source segment to be translated from English to Nko is prefixed as follows per model:
\\
206.19: ``<to\_nqo\_Nkoo> "
\\
207.19: ``<from\_eng\_Latn>~<to\_nqo\_Nkoo> "
\\
208.19: ``<from>~<eng\_Latn>~<to>~<nqo\_Nkoo> "
\\
209.19: ``<from><eng\_Latn><to><nqo\_Nkoo> "

\begin{table}[]
    \centering
    \begin{tabular}{ll}
        \toprule
         model &  prefix \\
         \midrule
        200.11 & (none) \\
        200.16 &  \\
        \hline
        201.16 & <to\_tgt\_Lang> \\
        202.16 &  \\
        206.19 &  \\
        \hline
        207.19 & <from\_src\_Lang> <to\_tgt\_Lang> \\
        \hline
        208.19 & <from> <src\_Lang> <to> <tgt\_Lang> \\
        \hline
        209.19 & <from\_src\_Lang\_to\_tgt\_Lang> \\
        \bottomrule
    \end{tabular}
    \caption{Specification of source sequence language token prefixes used in our multilingual translation models.}
    \label{tab:model-lang-tokens}
\end{table}

\iffalse
\begin{table}
    \centering
    \begin{tabular}{rrrrl}
        \toprule
        models & layers & drop & train & directions\\ 
        \midrule
        200.11 & 5+5 & 0.4 & 116,255 & nqo $\leftarrow$ eng \\ 
        \hline
        200.16 & 6+6 & 0.6 & 116,255 & nqo $\leftarrow$ eng \\ 
        \hline
        201.16 & 6+6 & 0.6 & 116,255 & nqo $\rightleftarrows$ eng \\
               &     &     & 71,408 & nqo $\rightleftarrows$ fra \\
               &     &     & 6,193 & nqo $\rightleftarrows$ bam \\
        \hline
        202.16 & 6+6 & 0.6 & 116,255 & nqo $\rightleftarrows$ eng \\
        206.19 &     &     & 71,408 & nqo $\rightleftarrows$ fra  \\
        207.19 &     &     & 6,193 & nqo $\rightleftarrows$ bam  \\
        208.19 &     &     & 325,842 & nqo $\leftarrow$ nqo \\
        209.19 &     &     &  &  \\
        \bottomrule
    \end{tabular}
    \caption{Models with their encoder/decoder layer counts (layers), input token embedding dropout probability (drop), and number of training segments per direction.}
\end{table}

\fi

\subsection{Training}
During training, dropout is used with the following probabilities: input token embedding dropout 0.4 (xxx.11) or 0.6 (xxx.16, xxx.19), attention dropout 0.2, ReLU dropout 0.2.
The label-smoothed cross-entropy loss function is used with a smoothing rate of 0.2.
Optimization is performed using Adam with a weight decay of 0.0001. 
The inverse squared root learning rate scheduler is used with an initial rate of 1e-7 and 4000 warm-up updates.
Gradient clipping is employed with a norm threshold of 1.
Effective batches of up to 65,536 tokens are used to train all models. Gradients are accumulated for 1 batch of up 65,536 tokens on A100 GPUs and 4 batches of up to 16384 on Titan XP GPUs before each update.

\subsection{Model Selection and Stopping Criteria}
Trainings are stopped when BLEU scores on the validation step do not improve after 20K gradient updates. Checkpoints with the highest BLEU scores on the validation set are selected. The average BLEU score across all supported translation directions is used for multilingual model selection.

\subsection{Evaluation}
CPC/FLoRes-dev and CPC/FLoRes-devtest are respectively used as validation and test sets. For each model, their subsets with languages of interest are considered (see Tables \ref{tab:data-details-valid-test} and \ref{tab:data-details-train} ).
The chrF++ score, which has been shown to align well with human assessments, especially for morphologically rich languages \cite{popovic2017chrf++}, is used as the main evaluation metric.
The Sacre BLEU library \cite{post-2018-call} is used to compute BLEU and chrF++  scores.

\subsection{Results}
Table \ref{tab:results} shows the test and validation BLEU and chrF++ scores for each model and supported translation direction. The best performing model 208.19 scores 26.00 mean chrF++ on the test set.

\paragraph{Layer Count and Regularization:}
Compared to 200.11, 200.16 with one extra encoder and decoder layer, and a higher token embedding dropout rate, scored $+0.34$ \emph{nqo} $\leftarrow$ \emph{eng} chrF++.
% 30.07 - 29.73 = 0.34

\paragraph{Multilinguality:}
Compared to 200.16, the multilingual model 201.16, which has the same architecture and training hyperparameters, scored $-0.92$ \emph{nqo} $\leftarrow$ \emph{eng} chrF++.
% 29.15 - 30.07 = -0.92

\paragraph{Monolingual Autoencoding:}
Compared to 201.16, 202.16, which performs simultaneous multilingual translation and monolingual autoencoding, scored $+0.14$ \emph{nqo} $\leftarrow$ \emph{eng} chrF++
% 24.26 - 24.12 = 0.14

\paragraph{Attention Heads and Shared Embeddings:}
Compared to 202.16, 206.19 which uses 8 attention heads in the encoder and decoder layers, and which shares all input and output embeddings and dictionaries scores $+1.59$ mean chrF++.
% 25.85 - 24.26 = 1.59

\paragraph{Language Token Prefixing:}
Compared to 206.19, which only specifies target language tokens in the source sequence, 207.19, which specifies the source and target languages as two separate tokens, scored $+0.08$ mean chrF++.
209.19, which specifies the source and target languages as a single token, scored $+0.06$ mean chrF++.
208.19, which specifies the source and target languages in a four-token clause scored $+0.15$ mean chrF++.

% 25.93 - 25.85 = 0.08
% 26 - 25.85 = 0.15
% 25.91 - 25.85 = 0.06

\begin{table}[]
    \centering
    \begin{adjustbox}{max width=0.48\textwidth}
    \begin{tabular}{@{}lccccc@{}}
            \toprule
        ~&~& \multicolumn{2}{c}{Intl. BLEU} & \multicolumn{2}{c}{chrF++} \\
        \cmidrule{3-4}  \cmidrule{5-6}
        model & direction & valid\ & test & valid & test \\
        \midrule
        200.11 & \emph{nqo $\leftarrow$ eng} & 5.40 & 5.11 & 28.80 & 29.73 \\
        \midrule
        200.16 & \emph{nqo $\leftarrow$ eng} & 5.85 & 5.25 & 29.06 & 30.07 \\
        \midrule
        201.16 & \emph{nqo $\rightarrow$ bam} & 1.19 & 1.12 & 16.73 & 17.04 \\
        201.16 & \emph{nqo $\leftarrow$ bam} & 2.86 & 3.19 & 22.07 & 23.01 \\
        201.16 & \emph{nqo $\rightarrow$ eng} & 3.65 & 3.78 & 26.31 & 26.99 \\
        201.16 & \emph{nqo $\leftarrow$ eng} & 6.11 & 5.71 & 28.64 & 29.15 \\
        201.16 & \emph{nqo $\rightarrow$ fra} & 2.33 & 2.35 & 22.27 & 22.61 \\
        201.16 & \emph{nqo $\leftarrow$ fra} & 4.50 & 4.29 & 25.55 & 25.89 \\
        201.16 & mean & 3.44 & 3.41 & 23.60 & 24.12 \\
        \midrule
        202.16 & \emph{nqo $\rightarrow$ bam} & 1.14 & 1.00 & 16.68 & 16.82 \\
        202.16 & \emph{nqo $\leftarrow$ bam} & 2.83 & 3.11 & 22.33 & 23.11 \\
        202.16 & \emph{nqo $\rightarrow$ eng} & 4.27 & 4.26 & 26.86 & 27.61 \\
        202.16 & \emph{nqo $\leftarrow$ eng} & 6.18 & 5.80 & 28.63 & 29.44 \\
        202.16 & \emph{nqo $\rightarrow$ fra} & 2.31 & 2.74 & 22.46 & 22.89 \\
        202.16 & \emph{nqo $\leftarrow$ fra} & 4.18 & 4.51 & 25.22 & 25.68 \\
        202.16 & mean & 3.49 & 3.57 & 23.70 & 24.26 \\
        \midrule
        206.19 & \emph{nqo $\rightarrow$ bam} & 1.69 & 1.50 & 19.04 & 19.34 \\
        206.19 & \emph{nqo $\leftarrow$ bam} & 3.63 & 3.43 & 23.26 & 23.81 \\
        206.19 & \emph{nqo $\rightarrow$ eng} & 5.22 & 5.15 & 28.70 & 28.85 \\
        206.19 & \emph{nqo $\leftarrow$ eng} & 6.79 & 6.50 & 29.97 & 30.66 \\
        206.19 & \emph{nqo $\rightarrow$ fra} & 3.28 & 3.41 & 25.26 & 25.42 \\
        206.19 & \emph{nqo $\leftarrow$ fra} & 4.77 & 5.05 & 26.59 & 27.03 \\
        206.19 & mean & 4.23 & 4.17 & 25.47 & 25.85 \\
        \midrule
        207.19 & \emph{nqo $\rightarrow$ bam} & 1.52 & 1.53 & 19.09 & \textbf{19.43} \\
        207.19 & \emph{nqo $\leftarrow$ bam} & 3.56 & 3.28 & 23.10 & 23.77 \\
        207.19 & \emph{nqo $\rightarrow$ eng} & 5.10 & 5.05 & 28.61 & 28.69 \\
        207.19 & \emph{nqo $\leftarrow$ eng} & 6.96 & 6.21 & 29.92 & 30.45 \\
        207.19 & \emph{nqo $\rightarrow$ fra} & 3.26 & 3.51 & 25.44 & \textbf{25.98} \\
        207.19 & \emph{nqo $\leftarrow$ fra} & 5.23 & 5.14 & 26.89 & \textbf{27.25} \\
        207.19 & mean & 4.27 & 4.12 & 25.51 & 25.93 \\
        \midrule
        208.19 & \emph{nqo $\rightarrow$ bam} & 1.44 & 1.52 & 18.83 & 19.08 \\
        208.19 & \emph{nqo $\leftarrow$ bam} & 3.32 & 3.37 & 23.38 & \textbf{24.00} \\
        208.19 & \emph{nqo $\rightarrow$ eng} & 4.78 & 5.05 & 28.64 & \textbf{29.13} \\
        208.19 & \emph{nqo $\leftarrow$ eng} & 6.99 & 6.44 & 30.05 & \textbf{30.83} \\
        208.19 & \emph{nqo $\rightarrow$ fra} & 3.20 & 3.61 & 25.15 & 25.79 \\
        208.19 & \emph{nqo $\leftarrow$ fra} & 5.04 & 4.78 & 26.73 & 27.17 \\
        208.19 & mean & 4.13 & 4.13 & 25.46 & \textbf{26.00} \\
        \midrule
        209.19 & \emph{nqo $\rightarrow$ bam} & 1.60 & 1.47 & 19.00 & 19.25 \\
        209.19 & \emph{nqo $\leftarrow$ bam} & 3.45 & 3.43 & 23.29 & 23.80 \\
        209.19 & \emph{nqo $\rightarrow$ eng} & 5.07 & 4.79 & 28.67 & 28.82 \\
        209.19 & \emph{nqo $\leftarrow$ eng} & 6.96 & 6.58 & 30.10 & 30.78 \\
        209.19 & \emph{nqo $\rightarrow$ fra} & 3.49 & 3.13 & 25.39 & 25.76 \\
        209.19 & \emph{nqo $\leftarrow$ fra} & 5.13 & 4.92 & 26.56 & 27.06 \\
        209.19 & mean & 4.28 & 4.05 & 25.50 & 25.91 \\
        \bottomrule
    \end{tabular}
    \end{adjustbox}
    \caption{Our bilingual and multilingual models measured for accuracy on FLoRes-dev (valid) and FLoRes-devtest (test) using the Intl. BLEU (Sacre BLEU with Unicode-aware tokenization) and chrF++ metrics.}
    \label{tab:results}
\end{table}

\section{Discussions}
\subsection{\systemname{}}

\paragraph{Improving Usability:}
As noted by Nko translators, the usability of \systemname{} could be improved by: (1) Allowing translators to review their recently submitted tasks before the Workflow Manager proceeds to the next stage of the curation process. (2) Implementing an offline authentication mechanism.

\paragraph{Adding a Translation Memory:}
Adding a translation memory could increase the productivity, accuracy, and consistency of translators. However, the effect of such a tool on the general quality of translations, including the diversity of synonyms and expression styles should not be overlooked.

\paragraph{Extensibility:}
Alternate copyediting workflows can be implemented in \systemname{} by extending the Workflow Manager and Task Manager. The task presentation user interface can also be adapted to other text curation tasks, such as syntax annotation.

\subsection{Parallel Corpora}
\paragraph{Handling Short Sequences:}
The segments in \emph{nicolingua-0005} are, on average, significantly shorter than those in FLoRes and NLLB-\textsc{Seed}.
Despite being short, sequences such as ones from the Nko-Français dictionary and Unicode CLDR files, are too valuable to discard.
To prevent biasing models towards shorter sequence lengths, we repeated the \emph{(nqo\_Nkoo, eng\_Latn)} data from CPC/NLLB-\textsc{Seed} five times in the training set. A more principled approach should be considered.

\paragraph{Punctuations, Case and Diacritical Marks:}
Our models showed sensitivity to minor changes in Latin case, and punctuation as well as Nko diacritical marks (see Appendix \ref{sec:generation-sensitivity-examples}). Including augmented data with lowered case and stripped punctuation and diacritical marks in source sequences in the training corpora may help address this issue.

\paragraph{Learning from Translator Edits:}
Translator edits, as recorded by \systemname{} throughout the copy-edit process, could be useful for various modeling and quality estimation tasks.
This data could also be used for an auxiliary copy-edit reconstruction task that may improve the accuracy of a multitask NMT model.
Finally, translator edit data can be used to train and align translators on consistency standards.

\subsection{Neural Machine Translation}

\paragraph{Tokenization:}
Subword regularization, as discussed in \cite{kudo-2018-subword-regularization} and the dropout-based approach presented by \cite{provilkov-etal-2020-bpe}, may lead to increased translation performance for Nko.

\paragraph{Language Token Prefixes:}
The choice of source-side prefixing strategy had a marginal impact on translation accuracy. 
Our best model employs a four-token prefix, consisting of source and target language tokens joined with the `<from>' and `<to>' tokens. Our results and those of \cite{wicks-duh-2022-effects}, suggest that the specification of translation directions as source-side prefixes in multilingual NMT models merits further investigation.

\paragraph{Learning from Monolingual Data:} The use of monolingual Nko data in 202.16 led to marginal improvements in most translation directions.  Additional unsupervised tasks such as masked language modeling and denoising should also be explored.

\paragraph{Data Augmentation:} Back-translation-based data augmentation, and the generalized data augmentation method in \cite{xia2019generalized} could significantly increase  NMT performance for Nko.

\paragraph{International BLEU}
Our BLEU scores are computed with sacreBLEU using international tokenization because sacreBLEU's current default tokenizer (v13a) is inappropriate for Nko; it doesn't properly interpret the Nko Unicode block, particularly its punctuations, to detect word boundaries.

\paragraph{BLEU vs chrF++}
The BLEU scores of our models are rather low. This was surprising given the training data size and given Nko translators' feedback on generated translations. This observation is in line with \cite{popovic2017chrf++}'s hypothesis that chrF++ correlates better with human judgment than BLEU for morphologically rich languages.

\section{Conclusion}
This work presented \systemname{}, a collaborative parallel text curation system with copyediting-based quality workflows. \systemname{} enabled the extension of existing multilingual corpora, FLoRes-200 and NLLB-\textsc{Seed} with high-quality Nko translations. Those, and a new corpus we introduced, \emph{nicolingua-0005}, served to build baseline bilingual and multilingual NMT systems for Nko, with the best model achieving the accuracy of $30.84$ \emph{eng\_Latn}~$\rightarrow$~\emph{nqo\_Nkoo} chrF++.
We have released \systemname{} to facilitate the development and extension of multilingual parallel corpora to more languages.
We have also released resources and tools to enable the reproducibility of our results, and further progress towards usable MT systems for Nko.

%In this paper we have presented the extension of FLoRes-200 and NLLB-\textsc{Seed} to Nko, emphasizing the resulting newly available data, our new software for creating parallel multilingual corpora, and improvements to the alignment of NLLB-\textsc{Seed}.

%In this paper, we presented baseline neural machine translation systems for translating to and from Nko, a West African language. We explored bilingual and multilingual Transformer-based models. The results on the FLoRes datasets remain modest, with the best model achieving 30.84 $eng\_Latn \rightarrow nqo\_Nkoo$ chrF++.
%The results of this study serve as a starting point for developing better machine translation systems for Nko.

\section*{Acknowledgements}
Sources of support that made this project possible include:
A generous unrestricted research gift from Meta Platforms, Inc., the Stanford Graduate Fellowship (SGF), the Stanford Natural Language Processing (NLP) group, FriaSoft, and Nko USA Inc.
We extend our deep gratitude to FriaSoft for selflessly donating numerous software engineering hours and to Nko USA Inc. for providing Nko language expertise and resources.
The commitment of Nko USA Inc. and FriaSoft to advancing West African language technology was instrumental in realizing this project.
We thanks to John Hewitt and Tol\'{u}l\d{o}p\d{\'{é}} \`{O}g\'{u}nr\d{\`{e}}m\'{i} for giving invaluable feedback.
We are grateful to Asmaou Diallo, Nantenin Camara, Koulako Camara, Moussa Thomas Doumbouya, and Ibrahima Doumbouya for creating a supportive environment conducive to the execution of portions of this work in Conakry, Fria, and Boké.

\bibliography{main}

\begin{thebibliography}{46}
\expandafter\ifx\csname natexlab\endcsname\relax\def\natexlab#1{#1}\fi

\bibitem[{Bryant(2020)}]{bryant2020education}
Kelly~Duke Bryant. 2020.
\newblock \href {https://oxfordre.com/africanhistory/display/10.1093/acrefore/9780190277734.001.0001/acrefore-9780190277734-e-679} {{E}ducation and {P}olitics in {C}olonial {F}rench {W}est {A}frica}.
\newblock In \emph{Oxford Research Encyclopedia of African History}.

\bibitem[{Burckhardt et~al.(2014)}]{burckhardt2014principles}
Sebastian Burckhardt et~al. 2014.
\newblock \href {https://www.nowpublishers.com/article/Details/PGL-011} {Principles of {E}ventual {C}onsistency}.
\newblock \emph{Foundations and Trends{\textregistered} in Programming Languages}, 1(1-2):1--150.

\bibitem[{Camara(1953)}]{camara_enfant_noir}
Laye Camara. 1953.
\newblock \emph{L'enfant {N}oir}.
\newblock Éditions Plon.

\bibitem[{Conde(2017)}]{conde2017}
Nafadji~Sory Conde. 2017.
\newblock \href {https://www.editions-harmattan.fr/livre-introduction_au_n_ko_une_alternative_linguistique_pour_l_afrique_nafadji_sory_conde-9782343119830-53315.html} {\emph{Introduction au N'ko: Une Alternative Linguistique pour l'Afrique}}.
\newblock Presses de l'Université Kofi Annan and Harmattan Guinée.

\bibitem[{Costa-juss{\`a} et~al.(2020)Costa-juss{\`a}, Creus, Domingo, Dom{\'\i}nguez, Escobar, L{\'o}pez, Garcia, and Geleta}]{costa2020mt}
Marta~R Costa-juss{\`a}, Roger Creus, Oriol Domingo, Albert Dom{\'\i}nguez, Miquel Escobar, Cayetana L{\'o}pez, Marina Garcia, and Margarita Geleta. 2020.
\newblock \href {https://arxiv.org/abs/2005.13156} {Mt-{A}dapted {D}atasheets for {D}atasets: {T}emplate and {R}epository}.
\newblock \emph{arXiv preprint arXiv:2005.13156}.

\bibitem[{Diane(2022)}]{kanjamadi_kanjamadi_2022}
Baba~Mamadi Diane. 2022.
\newblock \href {http://kanjamadi.org/baju/} {Kanjamadi – {Kanjamadi} for {Nko}}.
\newblock \url{https://web.archive.org/web/20231011145800/https://kanjamadi.org/baju/}.
\newblock Accessed on 2023-10-11.

\bibitem[{{Diane, Baba Mamadi}(2021)}]{quranenc:nqo}
{Diane, Baba Mamadi}. 2021.
\newblock Translation of the {M}eanings of the {N}oble {Q}ur'an - {N}'ko {T}ranslation.
\newblock \url{https://quranenc.com/en/browse/ankobambara_dayyan/1}.
\newblock Accessed on 2023-10-23.

\bibitem[{Donaldson(2017)}]{donaldson2017clear}
Coleman Donaldson. 2017.
\newblock \href {https://www.journals.uchicago.edu/doi/full/10.1086/702554} {\emph{Clear Language: Script, Register and the N'ko Movement of Manding-Speaking West Africa}}.
\newblock Ph.D. thesis, University of Pennsylvania, Philadelphia, PA.
\newblock Archived from the original on 2019-02-21. Retrieved 2019-02-21.

\bibitem[{Donaldson(2019)}]{Donaldson2019LinguisticAC}
Coleman Donaldson. 2019.
\newblock \href {https://api.semanticscholar.org/CorpusID:181625415} {Linguistic and {C}ivic {R}efinement in the {N}'ko {M}ovement of {M}anding-{S}peaking {W}est {A}frica}.
\newblock \emph{Signs and Society}, 7:156 -- 185.

\bibitem[{Doumbouya(2022)}]{Doumbouya_Detransliterator_2022}
Moussa Koulako~Bala Doumbouya. 2022.
\newblock \href {https://github.com/mdoumbouya/detransliterator} {{Detransliterator}}.
\newblock \url{https://github.com/mdoumbouya/detransliterator}.

\bibitem[{Eberhard et~al.(2023)Eberhard, Simons, and Fennig}]{ethnologue2023}
David~M. Eberhard, Gary~F. Simons, and Charles~D. Fennig. 2023.
\newblock \href {http://www.ethnologue.com} {Ethnologue: Languages of the world}.
\newblock Online version.

\bibitem[{Federmann(2018)}]{federmann-2018-appraise}
Christian Federmann. 2018.
\newblock \href {https://aclanthology.org/C18-2019} {Appraise {E}valuation {F}ramework for {M}achine {T}ranslation}.
\newblock In \emph{Proceedings of the 27th International Conference on Computational Linguistics: System Demonstrations}, pages 86--88, Santa Fe, New Mexico. Association for Computational Linguistics.

\bibitem[{Federmann et~al.(2022)Federmann, Kocmi, and Xin}]{federmann2022ntrex}
Christian Federmann, Tom Kocmi, and Ying Xin. 2022.
\newblock \href {https://aclanthology.org/2022.sumeval-1.4/} {{NTREX}-128--news {T}est {R}eferences for {MT} {E}valuation of 128 {L}anguages}.
\newblock In \emph{Proceedings of the First Workshop on Scaling Up Multilingual Evaluation}, pages 21--24.

\bibitem[{Google(2023)}]{firebase_offline}
Firebase~Documentation Google. 2023.
\newblock Access data offline.
\newblock \url{https://firebase.google.com/docs/firestore/manage-data/enable-offline}.
\newblock Accessed on 2023-08-04.

\bibitem[{{International Center, Noor}(2018)}]{quranenc:fra}
{International Center, Noor}. 2018.
\newblock Translation of the {M}eanings of the {N}oble {Q}ur'an - {F}rench {T}ranslation.
\newblock \url{https://quranenc.com/en/browse/french_montada/1}.
\newblock Accessed on 2023-08-15.

\bibitem[{{International, Saheeh}(2022)}]{quranenc:eng}
{International, Saheeh}. 2022.
\newblock Translation of the {M}eanings of the {N}oble {Q}ur'an - {E}nglish {T}ranslation.
\newblock \url{https://quranenc.com/en/browse/english_saheeh/1}.
\newblock Accessed on 2023-08-15.

\bibitem[{Johnson et~al.(2017)Johnson, Schuster, Le, Krikun, Wu, Chen, Thorat, Vi{\'e}gas, Wattenberg, Corrado et~al.}]{johnson2017google}
Melvin Johnson, Mike Schuster, Quoc~V Le, Maxim Krikun, Yonghui Wu, Zhifeng Chen, Nikhil Thorat, Fernanda Vi{\'e}gas, Martin Wattenberg, Greg Corrado, et~al. 2017.
\newblock \href {https://aclanthology.org/Q17-1024/} {Google’s {M}ultilingual {N}eural {M}achine {T}ranslation {S}ystem: {E}nabling {Z}ero-{S}hot {T}ranslation}.
\newblock \emph{Transactions of the Association for Computational Linguistics}, 5:339--351.

\bibitem[{Kotey(1975)}]{kotey1975official}
Paul~Amon Kotey. 1975.
\newblock The {O}fficial {L}anguage {C}ontroversy: {I}ndigenous versus {C}olonial.

\bibitem[{Kudo(2018)}]{kudo-2018-subword-regularization}
Taku Kudo. 2018.
\newblock \href {https://doi.org/10.18653/v1/P18-1007} {Subword {R}egularization: {I}mproving {N}eural {N}etwork {T}ranslation {M}odels with {M}ultiple {S}ubword {C}andidates}.
\newblock In \emph{Proceedings of the 56th Annual Meeting of the Association for Computational Linguistics (Volume 1: Long Papers)}, pages 66--75, Melbourne, Australia. Association for Computational Linguistics.

\bibitem[{Luong et~al.(2016)Luong, Le, Sutskever, Vinyals, and Kaiser}]{luong2016multi}
Minh-Thang Luong, Quoc~V. Le, Ilya Sutskever, Oriol Vinyals, and Lukasz Kaiser. 2016.
\newblock \href {https://www-nlp.stanford.edu/pubs/luong2016iclr_multi.pdf} {Multi-task {S}equence to {S}equence {L}earning}.
\newblock In \emph{International Conference on Learning Representations}.

\bibitem[{Mager et~al.(2021)Mager, Oncevay, Ebrahimi, Ortega, Gonzales, Fan, Gutierrez-Vasques, Chiruzzo, Gim{\'e}nez-Lugo, Ramos et~al.}]{mager2021findings}
Manuel Mager, Arturo Oncevay, Abteen Ebrahimi, John Ortega, Annette~Rios Gonzales, Angela Fan, Ximena Gutierrez-Vasques, Luis Chiruzzo, Gustavo Gim{\'e}nez-Lugo, Ricardo Ramos, et~al. 2021.
\newblock \href {https://aclanthology.org/2021.americasnlp-1.23/} {Findings of the {A}mericasnlp 2021 {S}hared {T}ask on {O}pen {M}achine {T}ranslation for {I}ndigenous {L}anguages of the {A}mericas}.
\newblock In \emph{Proceedings of the First Workshop on Natural Language Processing for Indigenous Languages of the Americas}, pages 202--217.

\bibitem[{Mozilla(2023{\natexlab{a}})}]{web_api_cache_storage}
MDN~Contributors Mozilla. 2023{\natexlab{a}}.
\newblock {CacheStorage} - {W}eb {API}s | {MDN}.
\newblock \url{https://developer.mozilla.org/en-US/docs/Web/API/CacheStorage}.
\newblock Accessed on 2023-08-04.

\bibitem[{Mozilla(2023{\natexlab{b}})}]{web_api_service_worker}
MDN~Contributors Mozilla. 2023{\natexlab{b}}.
\newblock Service {W}orker {API} - {MDN} {W}eb {D}ocs.
\newblock \url{https://developer.mozilla.org/en-US/docs/Web/API/Service_Worker_API}.
\newblock Accessed on 2023-08-04.

\bibitem[{Nekoto et~al.(2020)Nekoto, Marivate, Matsila, Fasubaa, Fagbohungbe, Akinola, Muhammad, Kabongo~Kabenamualu, Osei, Sackey, Niyongabo, Macharm, Ogayo, Ahia, Berhe, Adeyemi, Mokgesi-Selinga, Okegbemi, Martinus, Tajudeen, Degila, Ogueji, Siminyu, Kreutzer, Webster, Ali, Abbott, Orife, Ezeani, Dangana, Kamper, Elsahar, Duru, Kioko, Espoir, van Biljon, Whitenack, Onyefuluchi, Emezue, Dossou, Sibanda, Bassey, Olabiyi, Ramkilowan, {\"O}ktem, Akinfaderin, and Bashir}]{nekoto-etal-2020-participatory}
Wilhelmina Nekoto, Vukosi Marivate, Tshinondiwa Matsila, Timi Fasubaa, Taiwo Fagbohungbe, Solomon~Oluwole Akinola, Shamsuddeen Muhammad, Salomon Kabongo~Kabenamualu, Salomey Osei, Freshia Sackey, Rubungo~Andre Niyongabo, Ricky Macharm, Perez Ogayo, Orevaoghene Ahia, Musie~Meressa Berhe, Mofetoluwa Adeyemi, Masabata Mokgesi-Selinga, Lawrence Okegbemi, Laura Martinus, Kolawole Tajudeen, Kevin Degila, Kelechi Ogueji, Kathleen Siminyu, Julia Kreutzer, Jason Webster, Jamiil~Toure Ali, Jade Abbott, Iroro Orife, Ignatius Ezeani, Idris~Abdulkadir Dangana, Herman Kamper, Hady Elsahar, Goodness Duru, Ghollah Kioko, Murhabazi Espoir, Elan van Biljon, Daniel Whitenack, Christopher Onyefuluchi, Chris~Chinenye Emezue, Bonaventure F.~P. Dossou, Blessing Sibanda, Blessing Bassey, Ayodele Olabiyi, Arshath Ramkilowan, Alp {\"O}ktem, Adewale Akinfaderin, and Abdallah Bashir. 2020.
\newblock \href {https://doi.org/10.18653/v1/2020.findings-emnlp.195} {Participatory {R}esearch for {L}ow-{R}esourced {M}achine {T}ranslation: {A} {C}ase {S}tudy in {A}frican {L}anguages}.
\newblock In \emph{Findings of the {A}ssociation for {C}omputational {L}inguistics: EMNLP 2020}, pages 2144--2160, Online. Association for Computational Linguistics.

\bibitem[{Niane(1974)}]{niane1974histoire}
Djibril~Tamsir Niane. 1974.
\newblock Histoire et {T}radition {H}istorique du {M}anding.
\newblock \emph{Pr{\'e}sence africaine}, (1):59--74.

\bibitem[{{NLLB Team} et~al.(2022){NLLB Team}, Costa-jussà, Cross, Çelebi, Elbayad, Heafield, Heffernan, Kalbassi, Lam, Licht, Maillard, Sun, Wang, Wenzek, Youngblood, Akula, Barrault, Gonzalez, Hansanti, Hoffman, Jarrett, Sadagopan, Rowe, Spruit, Tran, Andrews, Ayan, Bhosale, Edunov, Fan, Gao, Goswami, Guzmán, Koehn, Mourachko, Ropers, Saleem, Schwenk, and Wang}]{nllb2022}
{NLLB Team}, Marta~R. Costa-jussà, James Cross, Onur Çelebi, Maha Elbayad, Kenneth Heafield, Kevin Heffernan, Elahe Kalbassi, Janice Lam, Daniel Licht, Jean Maillard, Anna Sun, Skyler Wang, Guillaume Wenzek, Al~Youngblood, Bapi Akula, Loic Barrault, Gabriel~Mejia Gonzalez, Prangthip Hansanti, John Hoffman, Semarley Jarrett, Kaushik~Ram Sadagopan, Dirk Rowe, Shannon Spruit, Chau Tran, Pierre Andrews, Necip~Fazil Ayan, Shruti Bhosale, Sergey Edunov, Angela Fan, Cynthia Gao, Vedanuj Goswami, Francisco Guzmán, Philipp Koehn, Alexandre Mourachko, Christophe Ropers, Safiyyah Saleem, Holger Schwenk, and Jeff Wang. 2022.
\newblock \href {https://arxiv.org/abs/2207.04672} {No {L}anguage {L}eft behind: {S}caling {H}uman-{C}entered {M}achine {T}ranslation}.

\bibitem[{Ott et~al.(2019)Ott, Edunov, Baevski, Fan, Gross, Ng, Grangier, and Auli}]{ott-etal-2019-fairseq}
Myle Ott, Sergey Edunov, Alexei Baevski, Angela Fan, Sam Gross, Nathan Ng, David Grangier, and Michael Auli. 2019.
\newblock \href {https://doi.org/10.18653/v1/N19-4009} {fairseq: {A} {F}ast, {E}xtensible {T}oolkit for {S}equence {M}odeling}.
\newblock In \emph{Proceedings of the 2019 Conference of the North {A}merican Chapter of the Association for Computational Linguistics (Demonstrations)}, pages 48--53, Minneapolis, Minnesota. Association for Computational Linguistics.

\bibitem[{Oyler(2002)}]{oyler2002}
Dianne Oyler. 2002.
\newblock \href {https://doi.org/10.1353/ral.2002.0034} {Re-{I}nventing {O}ral {T}radition: {T}he {M}odern {E}pic of {S}ouleymane {K}anté}.
\newblock \emph{Research in African Literatures}, 33(1):75--93.

\bibitem[{Pamula et~al.(2014)Pamula, Jairam, and Rajesh}]{pamula2014cache}
Narendra~Babu Pamula, K~Jairam, and B~Rajesh. 2014.
\newblock \href {https://citeseerx.ist.psu.edu/document?repid=rep1&type=pdf&doi=9fa410192ebccc36fdfdfd5bfdf0df5a1f60b583} {Cache-aside {A}pproach for {C}loud {D}esign {P}attern}.
\newblock \emph{International Journal of Computer Science and Information Technologies}, 5(2):1423--1426.

\bibitem[{Popovi{\'c}(2017)}]{popovic2017chrf++}
Maja Popovi{\'c}. 2017.
\newblock \href {https://aclanthology.org/W17-4770/} {chrf++: {W}ords {H}elping {C}haracter n-grams}.
\newblock In \emph{Proceedings of the second conference on machine translation}, pages 612--618.

\bibitem[{Post(2018)}]{post-2018-call}
Matt Post. 2018.
\newblock \href {https://doi.org/10.18653/v1/W18-6319} {A {C}all for {C}larity in {R}eporting {BLEU} {S}cores}.
\newblock In \emph{Proceedings of the Third Conference on Machine Translation: Research Papers}, pages 186--191, Brussels, Belgium. Association for Computational Linguistics.

\bibitem[{Provilkov et~al.(2020)Provilkov, Emelianenko, and Voita}]{provilkov-etal-2020-bpe}
Ivan Provilkov, Dmitrii Emelianenko, and Elena Voita. 2020.
\newblock \href {https://doi.org/10.18653/v1/2020.acl-main.170} {{BPE}-dropout: {S}imple and {E}ffective {S}ubword {R}egularization}.
\newblock In \emph{Proceedings of the 58th Annual Meeting of the Association for Computational Linguistics}, pages 1882--1892, Online. Association for Computational Linguistics.

\bibitem[{RFI(2016)}]{rfi_mandenkan_2016}
RFI. 2016.
\newblock \href {https://www.rfi.fr/fr/afrique/20161018-mandenkan-vitalite-une-langue} {Mandenkan, la {V}italité d'une {L}angue}.
\newblock {\url{https://www.rfi.fr/fr/afrique/20161018-mandenkan-vitalite-une-langue}}.
\newblock Accessed on 2023-10-11.

\bibitem[{Rosenberg(2011)}]{rosenberg_everyone_2011}
Tina Rosenberg. 2011.
\newblock \href {https://www.nytimes.com/2011/12/11/magazine/everyone-speaks-text-message.html} {Everyone {Speaks} {Text} {Message} - {The} {New} {York} {Times}}.
\newblock {\url{https://www.nytimes.com/2011/12/11/magazine/everyone-speaks-text-message.html}}.
\newblock Accessed on 2023-10-23].

\bibitem[{Sennrich et~al.(2016)Sennrich, Haddow, and Birch}]{sennrich-etal-2016-neural}
Rico Sennrich, Barry Haddow, and Alexandra Birch. 2016.
\newblock \href {https://doi.org/10.18653/v1/P16-1162} {Neural {M}achine {T}ranslation of {R}are {W}ords with {S}ubword {U}nits}.
\newblock In \emph{Proceedings of the 54th Annual Meeting of the Association for Computational Linguistics (Volume 1: Long Papers)}, pages 1715--1725, Berlin, Germany. Association for Computational Linguistics.

\bibitem[{{SIL International \& United Bible Societies}(2023)}]{paratext}
{SIL International \& United Bible Societies}. 2023.
\newblock \href {https://paratext.org/} {Paratext}.
\newblock \url{https://paratext.org/}.
\newblock Accessed on 2023-10-17.

\bibitem[{Soh et~al.(2021)Soh, Del~Carpio, and Wang}]{soh_impact_2021}
Yew~Chong Soh, Ximena~V. Del~Carpio, and Liang~Choon Wang. 2021.
\newblock \href {https://doi.org/10.1596/1813-9450-9517} {\emph{The {Impact} of {Language} of {Instruction} in {Schools} on {Student} {Achievement}: {Evidence} from {Malaysia} using the {Synthetic} {Control} {Method}}}.
\newblock Policy {Research} {Working} {Papers}. The World Bank.

\bibitem[{Unicode(2023{\natexlab{a}})}]{unicode_cldr}
Unicode. 2023{\natexlab{a}}.
\newblock \href {https://cldr.unicode.org/} {Unicode {CLDR} {P}roject}.
\newblock {\url{https://cldr.unicode.org/}}.
\newblock Accessed on 2023-06-15.

\bibitem[{Unicode(2023{\natexlab{b}})}]{unicode_cldr_acknowledgements}
Unicode. 2023{\natexlab{b}}.
\newblock \href {https://cldr.unicode.org/index/acknowledgments} {Unicode cldr project - acknowledgments}.
\newblock {\url{https://cldr.unicode.org/index/acknowledgments}}.
\newblock Accessed on 2023-06-15.

\bibitem[{Vaswani et~al.(2017)Vaswani, Shazeer, Parmar, Uszkoreit, Jones, Gomez, Kaiser, and Polosukhin}]{vaswani2017attention}
Ashish Vaswani, Noam Shazeer, Niki Parmar, Jakob Uszkoreit, Llion Jones, Aidan~N Gomez, {\L}ukasz Kaiser, and Illia Polosukhin. 2017.
\newblock \href {https://papers.nips.cc/paper_files/paper/2017/hash/3f5ee243547dee91fbd053c1c4a845aa-Abstract.html} {Attention is {A}ll you {N}eed}.
\newblock \emph{Advances in neural information processing systems}, 30.

\bibitem[{Virtanen et~al.(2020)Virtanen, Gommers, Oliphant, Haberland, Reddy, Cournapeau, Burovski, Peterson, Weckesser, Bright, {van der Walt}, Brett, Wilson, Millman, Mayorov, Nelson, Jones, Kern, Larson, Carey, Polat, Feng, Moore, {VanderPlas}, Laxalde, Perktold, Cimrman, Henriksen, Quintero, Harris, Archibald, Ribeiro, Pedregosa, {van Mulbregt}, and {SciPy 1.0 Contributors}}]{2020SciPy-NMeth}
Pauli Virtanen, Ralf Gommers, Travis~E. Oliphant, Matt Haberland, Tyler Reddy, David Cournapeau, Evgeni Burovski, Pearu Peterson, Warren Weckesser, Jonathan Bright, St{\'e}fan~J. {van der Walt}, Matthew Brett, Joshua Wilson, K.~Jarrod Millman, Nikolay Mayorov, Andrew R.~J. Nelson, Eric Jones, Robert Kern, Eric Larson, C~J Carey, {\.I}lhan Polat, Yu~Feng, Eric~W. Moore, Jake {VanderPlas}, Denis Laxalde, Josef Perktold, Robert Cimrman, Ian Henriksen, E.~A. Quintero, Charles~R. Harris, Anne~M. Archibald, Ant{\^o}nio~H. Ribeiro, Fabian Pedregosa, Paul {van Mulbregt}, and {SciPy 1.0 Contributors}. 2020.
\newblock \href {https://doi.org/10.1038/s41592-019-0686-2} {{{SciPy} 1.0: {F}undamental {A}lgorithms for {S}cientific {C}omputing in {P}ython}}.
\newblock \emph{Nature Methods}, 17:261--272.

\bibitem[{Vydrin et~al.(2016)Vydrin, Rovenchak, and Maslinsky}]{vydrin2016maninka}
Valentin Vydrin, Andrij Rovenchak, and Kirill Maslinsky. 2016.
\newblock \href {https://shs.hal.science/halshs-01358144/document} {Maninka {R}eference {C}orpus: {A} {P}resentation}.
\newblock In \emph{TALAf 2016: Traitement Automatique des Langues Africaines ({\'E}crit et Parole). Atelier JEP-TALN-RECITAL 2016-Paris le}.

\bibitem[{w3.org(2022)}]{w3_service_workers}
w3.org. 2022.
\newblock Service {W}orkers.
\newblock \url{https://www.w3.org/TR/service-workers/}.
\newblock Accessed on 2023-08-04.

\bibitem[{Wicks and Duh(2022)}]{wicks-duh-2022-effects}
Rachel Wicks and Kevin Duh. 2022.
\newblock \href {https://aclanthology.org/2022.aacl-short.19} {The {E}ffects of {L}anguage {T}oken {P}refixing for {M}ultilingual {M}achine {T}ranslation}.
\newblock In \emph{Proceedings of the 2nd Conference of the Asia-Pacific Chapter of the Association for Computational Linguistics and the 12th International Joint Conference on Natural Language Processing (Volume 2: Short Papers)}, pages 148--153, Online only. Association for Computational Linguistics.

\bibitem[{{Wikimedia}(2023)}]{listofwikipedias}
{Wikimedia}. 2023.
\newblock \href {https://meta.wikimedia.org/wiki/List_of_Wikipedias} {List of {W}ikipedias}.
\newblock {\url{https://meta.wikimedia.org/wiki/List_of_Wikipedias}}.
\newblock Accessed on 2023-10-14.

\bibitem[{Xia et~al.(2019)Xia, Kong, Anastasopoulos, and Neubig}]{xia2019generalized}
Mengzhou Xia, Xiang Kong, Antonios Anastasopoulos, and Graham Neubig. 2019.
\newblock \href {https://aclanthology.org/P19-1579/} {Generalized {D}ata {A}ugmentation for {L}ow-{R}esource {T}ranslation}.
\newblock In \emph{Proceedings of the 57th Annual Meeting of the Association for Computational Linguistics}, volume~57.

\end{thebibliography}
\bibliographystyle{acl_natbib}

%%%%%%%%%%%%%%%%%
% Friallel Appendices
%%%%%%%%%%%%%%%%%
\clearpage
\section*{Appendices}
\appendix

\section{\systemname{} User Study Feedback Questionnaire}
\begin{figure*}[tp]
    \centering
    \includegraphics[width=0.95\textwidth,trim={2.5cm 10cm 2.5cm 2cm},clip]{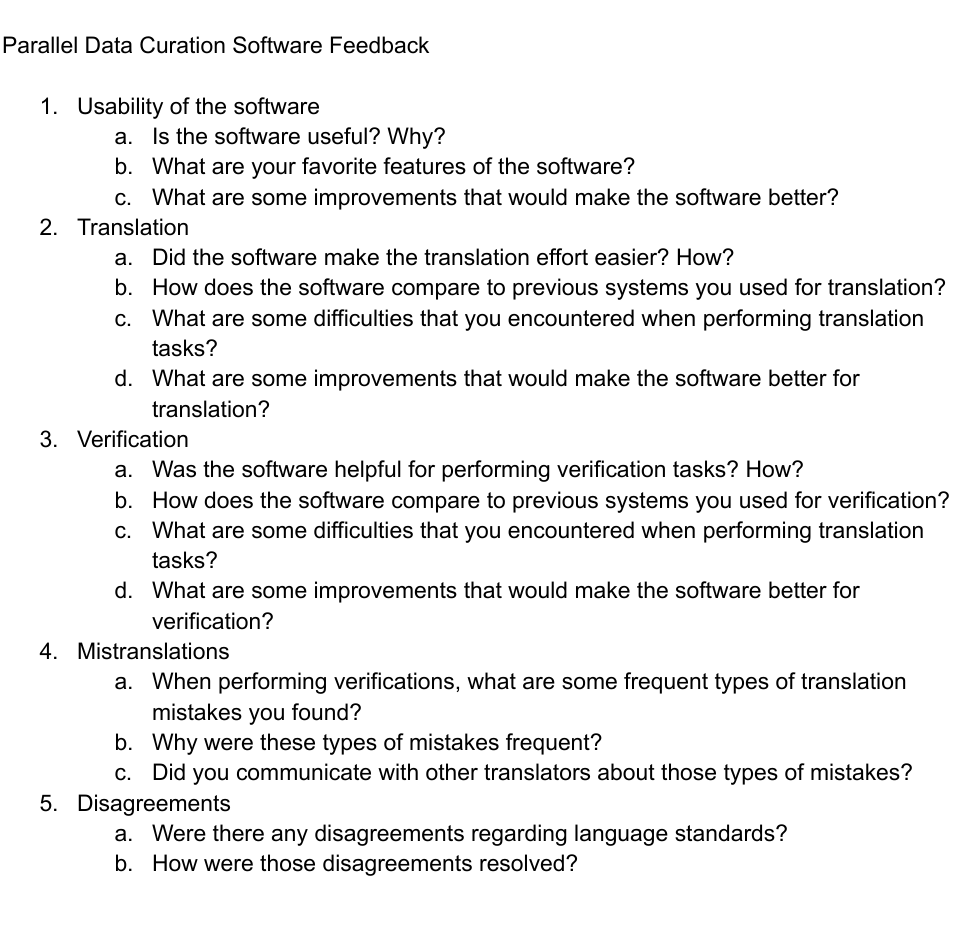}
    \caption{Survey questions sent to translators after they translated flores-200, nllb-seed, and ntrex-128 to N'Ko}
    \label{fig:survwy-questionnaire}
\end{figure*}

The feedback questionnaire sent to N'Ko translators appears in Figure~\ref{fig:survwy-questionnaire}.
% on the next page.

\clearpage

\section{\systemname{} Software Engineering Diagrams}
On the next three pages appear:
\begin{itemize}
    \item Workflow and task management sequence diagrams
    \item Workflow and Task State-Transition Diagrams
    \item Logical Data Model
    \item Physical Data Storage Model in Google Firestore
\end{itemize}

\begin{figure*}
    \centering
    \includegraphics[width=0.4\textwidth]{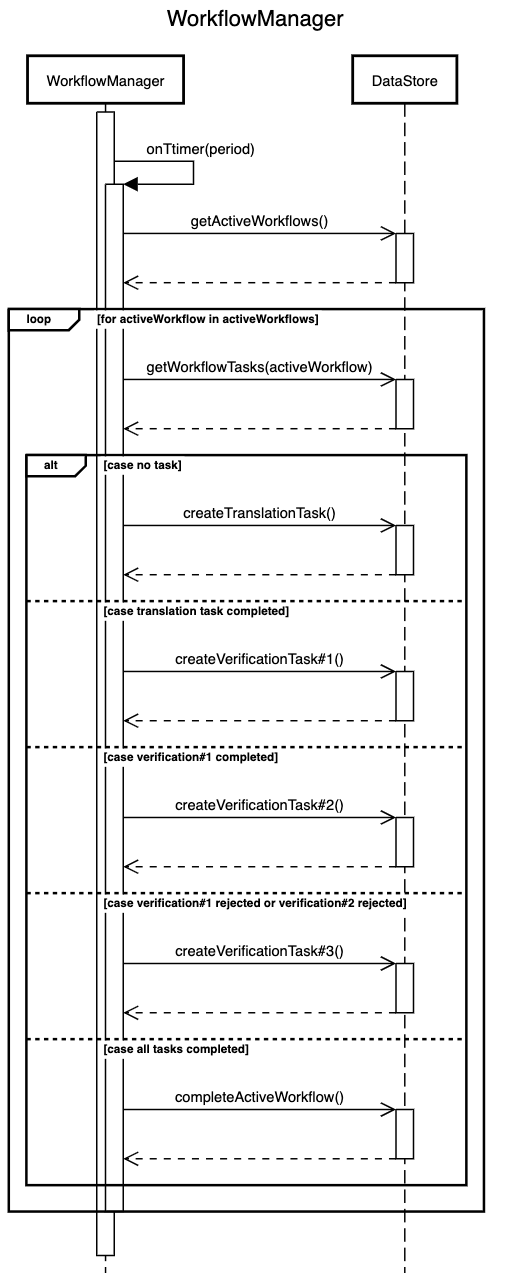}
    \includegraphics[width=0.5\textwidth]{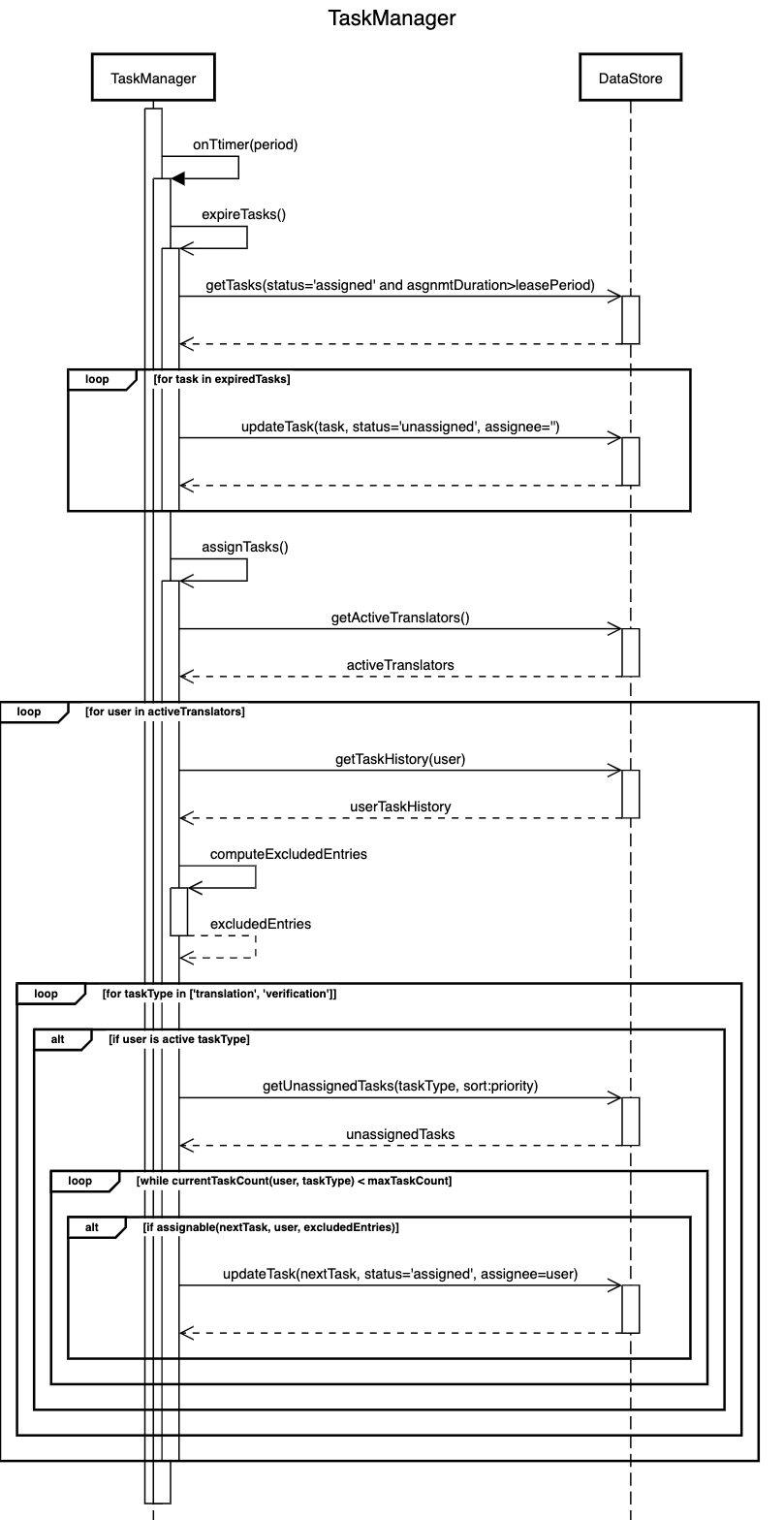}
    \caption{Sequence Diagram: Workflow Manager and Task Manager}
    \label{fig:sequence-diag-workflow-manager}
\end{figure*}

\begin{figure*}
    \centering
    \includegraphics[width=0.7\textwidth]{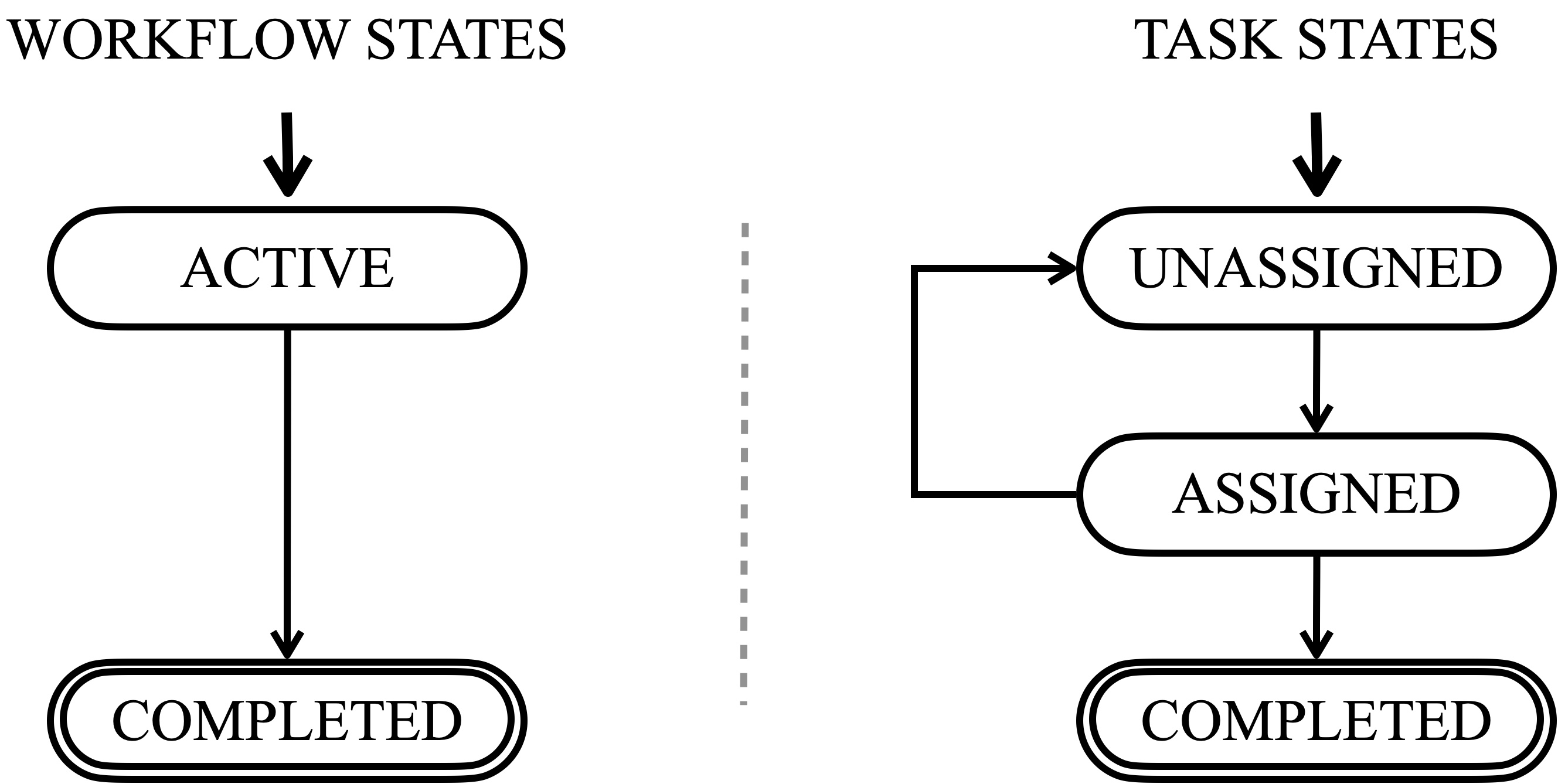}
    \caption{State Transition Diagrams for Tasks and Workflows. A translation workflow entity in the \emph{active} state is created for each dataset entry. The workflow manager creates related \emph{unassigned} tasks as needed, per the rules of the workflow. The Task Manager assigns tasks to users as appropriate. Uncompleted tasks are moved back to the \emph{unassigned} status when not completed within the lease period. The workflow manager moves workflows to the \emph{completed} status when all related tasks are \emph{completed} and there is no need to create additional tasks.}
    \label{fig:workflow}
\end{figure*}

\begin{figure*}
    \centering
    \includegraphics[width=0.95\textwidth]{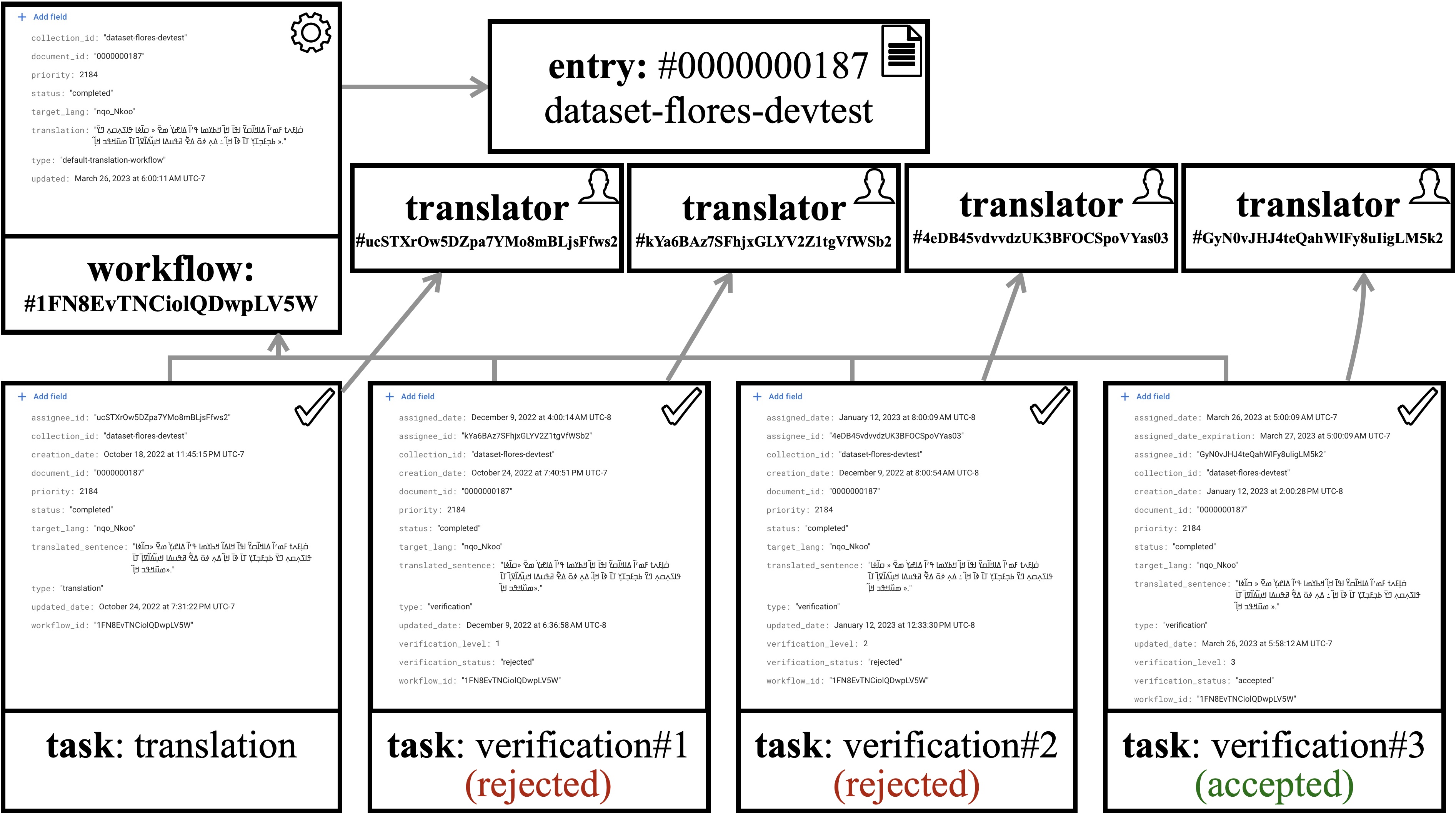}
    \caption{Logical model of entities involved in the curation process of entry\#187 of the FLoRes-devtest dataset. Each entity is stored as a document in the Firestore database. The Workflow Manager created one translation task, and three verification tasks, each assigned to a different translator. The third verification task was created because at least one of the previous two resulted in translator edits. Arrows point from referencing to referenced documents.}
    \label{fig:logical-model-workflows-and-tasks}
\end{figure*}

\begin{figure*}
    \centering
    \includegraphics[width=0.95\textwidth]{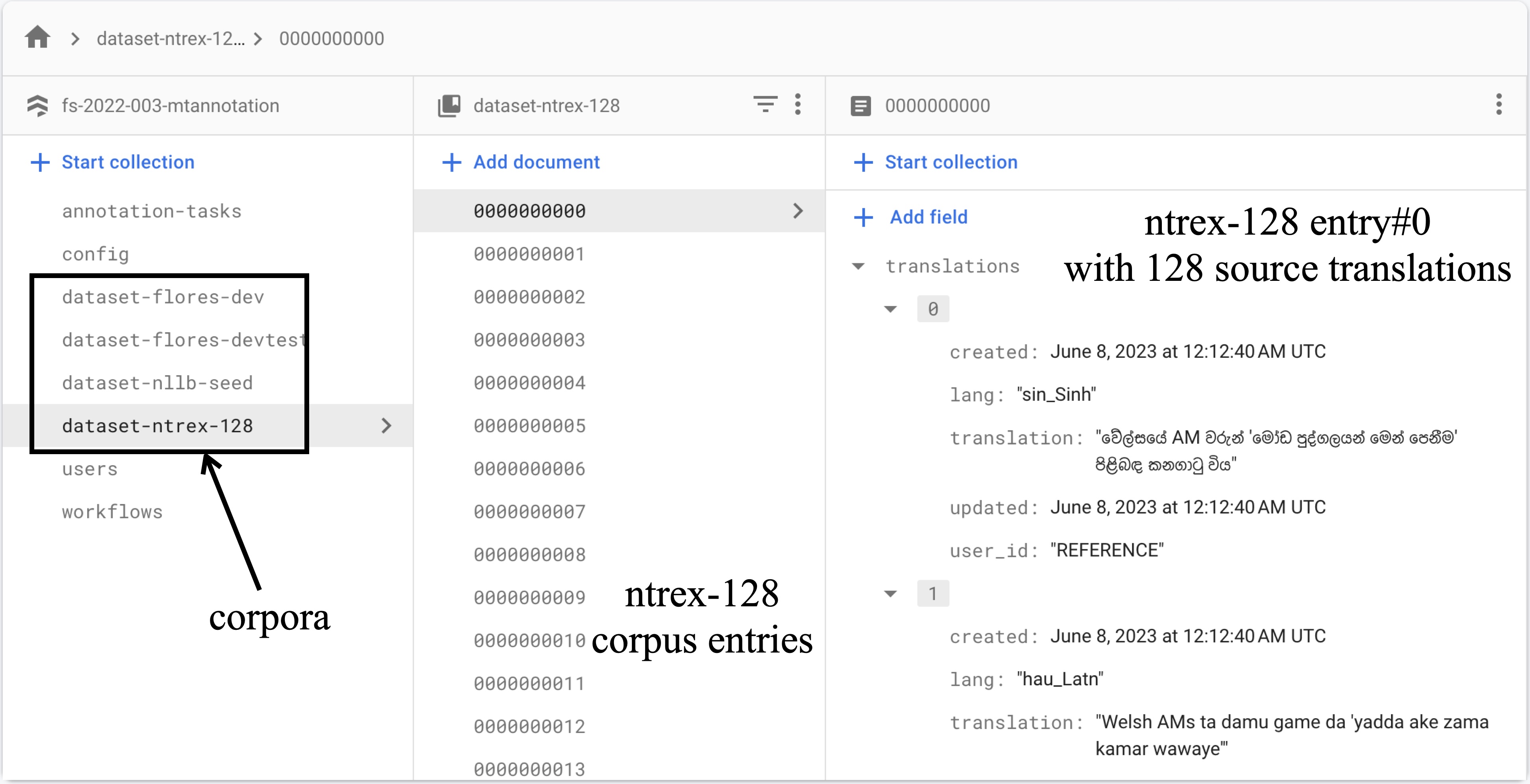}
    \caption{Data Storage in Google Firestore. Each corpus is stored as a collection of documents (left), each of which is identified by its position in the original data files (middle). Each entry contains an array of source translations. Each translation is labeled with its language and script codes (ISO-639\_ISO-15924) (right). The system also uses the \emph{users}, \emph{config}, \emph{workflows} and \emph{annotation-tasks} for user, configurations, and data curation workflow  management.}
    \label{fig:corpora-storage}
\end{figure*}

\iffalse

\begin{figure*}
    \centering
    \includegraphics[width=0.6\textwidth]{img/user-document.jpg}
    \caption{Example of user document. The user roles  \emph{isActiveTranslator} and \emph{isActiveVerifier} specify the types of tasks the user can perform, respectively, translation tasks, and verification/editing tasks. \emph{translation\_from\_languages} specify the languages this user can translate from, and which will be displayed if available in the source corpus.}
    \label{fig:user-document}
\end{figure*}

\fi

%%%%%%%%%%%%%%%%%
% CPC Appendices
%%%%%%%%%%%%%%%%%

\clearpage

\section{NLLB-\textsc{Seed} \emph{bam\_Latn} Quality Issues}
Examples of quality issues in NLLB-\textsc{Seed} \emph{bam\_Latn} data file 
appear on the following page.

\begin{figure*}
    \centering
    \includegraphics[width=0.8\linewidth]{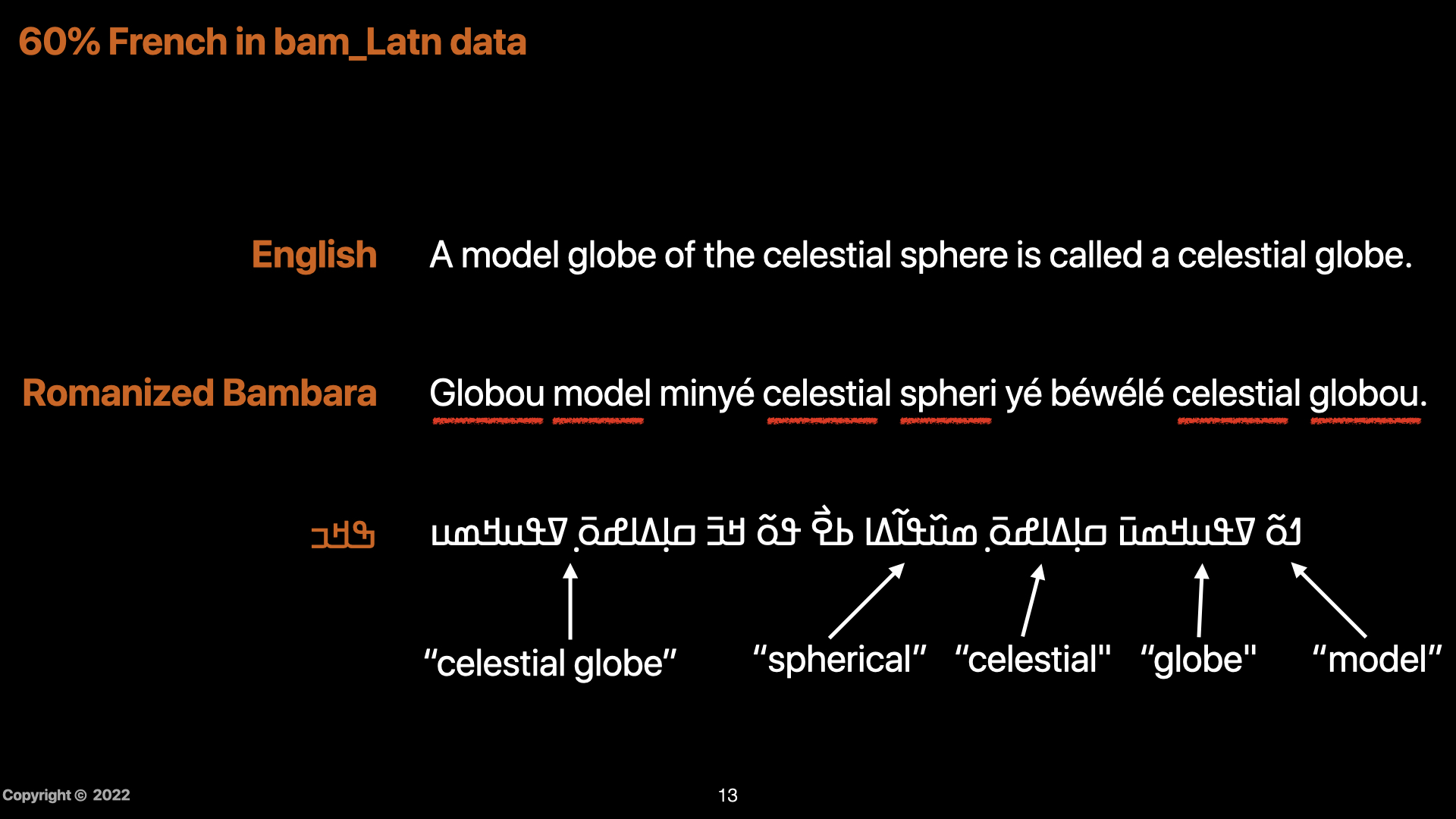}
    
    \includegraphics[width=0.8\linewidth]{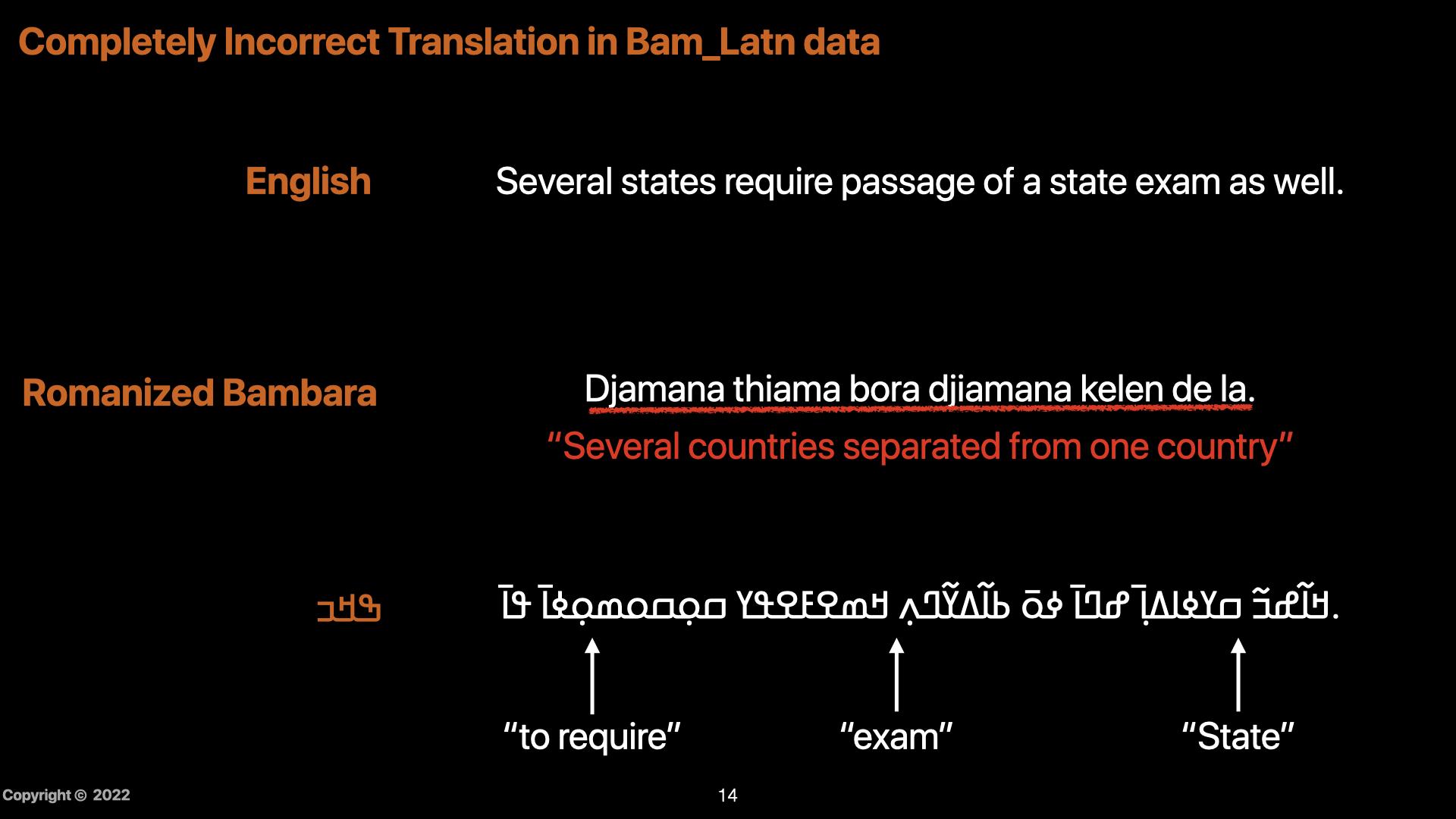}
    
    \includegraphics[width=0.8\linewidth]{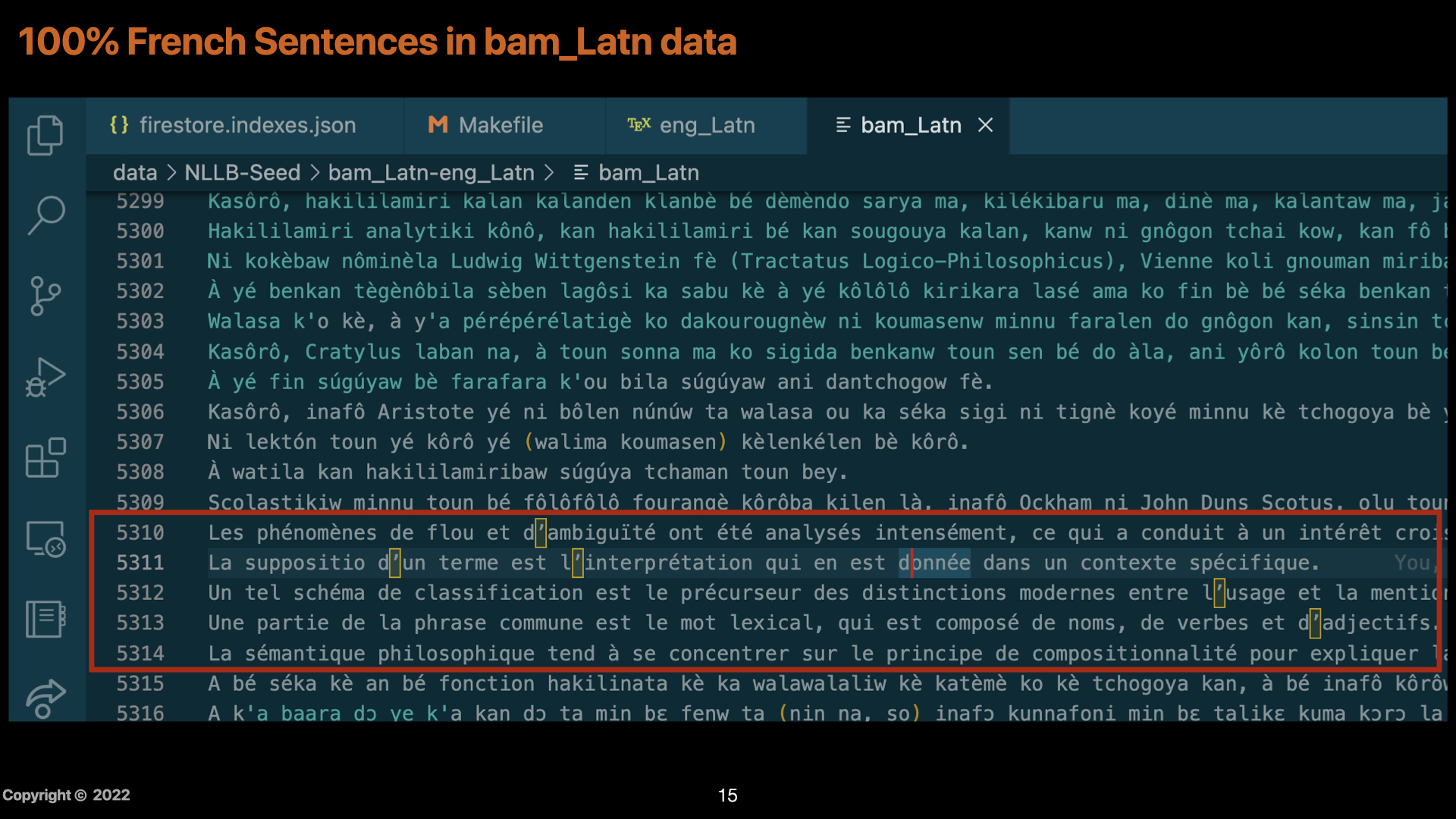}
    
    \caption{Examples of quality issues in NLLB-\textsc{Seed} \emph{bam\_Latn} data file. (1) Sentence with 60\% borrowed French words. (2) Incorrect translation. (3) a block of sentences entirely in French. Notice that tonal marks are missing from \emph{bam\_Latn} text.}
    
    \label{fig:bam_latn_quality_issues}
\end{figure*}

\clearpage

%%%%%%%%%%%%%%%%%
% CPC Appendices
%%%%%%%%%%%%%%%%%

\section{nicolingua-0005 details}
\label{sec::nicolingua-0005-details}

This section provides details on the monolingual, bilingual and trilingual parallel corpora, donated by Nko community members,  collectively making up the nicolingua-0005 corpus.

\subsubsection{Trilingual Corpora $(nqo\_Nkoo, eng\_Latn, fra\_Latn)$}

\paragraph{baba\_mamadi\_diane\_parallel\_002}
This corpus is composed of parallel Quran translations in Nko \cite{quranenc:nqo}, English \cite{quranenc:eng}, and French \cite{quranenc:fra}. The Quran's translation in Nko was originally performed by Baba Mamadi Diane for Islamic education purposes.

\paragraph{kalo\_mory\_diane\_parallel\_00\{1,2,3\}} This corpus contains various short phrases collected and translated by Kalo Mory Diane for the purpose of machine translation system development.

\paragraph{solo\_farabado\_cisse\_parallel\_002} This corpus contains various short phrases collected and translated by Solo Farabado Cisse for the purpose of machine translation system development.

\paragraph{solo\_farabado\_cisse\_parallel\_001} Nko localization strings from the Unicode Common Locale Data Repository (CLDR) \cite{unicode_cldr} to which Solo Farabado Cisse and Baba Mamadi Diane contributed \cite{unicode_cldr_acknowledgements}. Corresponding CLDR strings in Nko, English, and French were compiled to make this trilingual parallel corpus.

\subsubsection{Bilingual Corpora $(nqo\_Nkoo, eng\_Latn)$ }

\paragraph{baba\_mamadi\_diane\_parallel\_003} 
This corpus contains segments manually chunked from the Quran and translated by Baba Mamadi Diane specifically for the purpose of creating a corpus usable for machine translation system development.

\paragraph{baba\_mamadi\_diane\_parallel\_004} 
This corpus contains the localization strings of a custom Android 
 build translated by Baba Mamadi Diane.

\paragraph{djibrila\_diane\_parallel\_003} 
This corpus contains short phrases collected and translated by Djibrila Diane. The phrases also include some basic scientific terminology.
The corpus was originally created for education purposes only.

\paragraph{djibrila\_diane\_parallel\_001} 
This corpus contains short phrases in various tenses collected and translated by Djibrila Diane to serve of MT system development.

\paragraph{djibrila\_diane\_parallel\_002} 
This corpus contains various short phrases composed and translated by Djibrila Diane for the purpose of MT system development.

\subsubsection{Bilingual Corpora $(nqo\_Nkoo, fra\_Latn)$}

\paragraph{baba\_mamadi\_diane\_parallel\_001} Nko-French dictionary authored by Baba Mamadi Diane for education purposes. Dictionary entries in french with multiple forms (e.g. gender) were automatically expanded using regular expressions.

\paragraph{nafadji\_sory\_conde\_parallel\_001} 
This corpus contains various short phrases composed and translated by Nafadji Sory Conde for the purpose of machine translation system development.

\paragraph{nafadji\_sory\_conde\_parallel\_003} 
This corpus contains phrases from Camara Laye's 1953 novel ``L'enfant Noir'' \cite{camara_enfant_noir}.
The translation was originally done by Nafadji Sory Conde for the purpose of expanding available literature in Nko.

\paragraph{nafadji\_sory\_conde\_parallel\_002}
This corpus contains various phrases related to Guinean society and sociology. It was created by Nafadji Sory Conde for the purpose of MT system development.

\paragraph{nafadji\_sory\_conde\_parallel\_004} 
This corpus contains segments extracted from the Guinean constitution. It was originally translated by Nafaji Sory Conde for education purposes.

\subsubsection{Monolingual Corpora $(nqo\_Nkoo)$ }

\paragraph{nafadji\_sory\_conde\_monolingual\_001} 
This corpus, composed by Nafadji Sory Conde and his collaborators, contains extracts of books and newspapers in Nko.
A substantial part of the corpus was harvested from Kanjamadi.com. This corpus may overlap with the Maninka Reference Corpus \cite{vydrin2016maninka}.

\paragraph{baba\_mamadi\_diane\_monolingual\_00\{1,2\}}
These corpora were extracted from various Nko books and articles in various domains including history, religion, philosophy, literature and Science.
The corpora were originally composed by Baba M. Diane for the purpose of auto-completion algorithm development for Nko.

%%%%%%%%%%%%%%%%%%%%%%%%%%%%%%%%%

\begin{table*}
    \centering
    \begin{adjustbox}{max width=0.95\textwidth}
    \begin{tabular}{rrlll}
    \toprule
        lines & words & file & originator & description\\
    \midrule
        6236 & 175382 & baba\_mamadi\_diane\_parallel\_002.nqo\_Nkoo 
        & \multirow{3}{*}{Baba Mamadi Diane}
        & \multirow{3}{*}{\shortstack{Traductions of the Quran}} \\
        
        6236 & 151323 & baba\_mamadi\_diane\_parallel\_002.eng\_Latn\\
        6236 & 171085 & baba\_mamadi\_diane\_parallel\_002.fra\_Latn\\
        \hline
        7001 & 28626 & kalo\_mory\_diane\_parallel\_001.nqo\_Nkoo
        & \multirow{3}{*}{Kalo Mory Diane}
        & \multirow{3}{*}{\shortstack{Short Phrases}} \\
        
        7001 & 17558 & kalo\_mory\_diane\_parallel\_001.eng\_Latn\\
        7001 & 21593 & kalo\_mory\_diane\_parallel\_001.fra\_Latn\\
        \hline
        4001 & 18864 & kalo\_mory\_diane\_parallel\_003.nqo\_Nkoo
        & \multirow{3}{*}{Kalo Mory Diane}
        & \multirow{3}{*}{\shortstack{Short Phrases}} \\
        4001 & 12891 & kalo\_mory\_diane\_parallel\_003.eng\_Latn\\
        4001 & 15050 & kalo\_mory\_diane\_parallel\_003.fra\_Latn\\
        \hline
        3999 & 17903 & kalo\_mory\_diane\_parallel\_002.nqo\_Nkoo
        & \multirow{3}{*}{Kalo Mory Diane}
        & \multirow{3}{*}{\shortstack{Short Phrases}} \\
        3999 & 12237 & kalo\_mory\_diane\_parallel\_002.eng\_Latn\\
        3999 & 14495 & kalo\_mory\_diane\_parallel\_002.fra\_Latn\\
        \hline
        3052 & 13420 & solo\_farabado\_cisse\_parallel\_002.nqo\_Nkoo
        & \multirow{3}{*}{Solo Farabado Cisse}
        & \multirow{3}{*}{\shortstack{Short Phrases}} \\
        3052 & 9615 & solo\_farabado\_cisse\_parallel\_002.eng\_Latn\\
        3052 & 11308 & solo\_farabado\_cisse\_parallel\_002.fra\_Latn\\
        \hline
        1559 & 2739 & solo\_farabado\_cisse\_parallel\_001.nqo\_Nkoo
        & \multirow{3}{*}{Solo Farabado Cisse}
        & \multirow{3}{*}{\shortstack{Unicode CLDR Strings}} \\
        1559 & 2382 & solo\_farabado\_cisse\_parallel\_001.eng\_Latn\\
        1559 & 2338 & solo\_farabado\_cisse\_parallel\_001.fra\_Latn\\
        \bottomrule
    \end{tabular}
    \end{adjustbox}
    \caption{nicolingua-0005's trilingual subsets in Nko (\emph{nqo\_Nkoo}), English (\emph{eng\_Latn}) and French (\emph{fra\_Latn}) }
    \label{tab:nicolingua-0005-trilingual-corpora}
\end{table*}

\begin{table*}
    \centering
    \begin{adjustbox}{max width=0.95\textwidth}
    \begin{tabular}{rrlll}
    \toprule
        lines & words & file & originator & description\\
    \midrule
        21590 & 154238 & baba\_mamadi\_diane\_parallel\_003.nqo\_Nkoo
        & \multirow{2}{*}{Baba Mamadi Diane}
        & \multirow{2}{*}{\shortstack{Segments Chunked from the Quran}} \\
        21590 & 133369 & baba\_mamadi\_diane\_parallel\_003.eng\_Latn \\
        \hline
        36211 & 119536 & baba\_mamadi\_diane\_parallel\_004.nqo\_Nkoo
        & \multirow{2}{*}{Baba Mamadi Diane}
        & \multirow{2}{*}{\shortstack{Localization Strings for \\a Custom Android Build}} \\
        36211 & 72612 & baba\_mamadi\_diane\_parallel\_004.eng\_Latn \\
        \hline
        492 & 4666 & djibrila\_diane\_parallel\_003.nqo\_Nkoo
        & \multirow{2}{*}{Djibrila Diane}
        & \multirow{2}{*}{\shortstack{Various Short Phrases \\ and Basic Sci. Terms}} \\
        492 & 4122 & djibrila\_diane\_parallel\_003.eng\_Latn \\
        \hline
        1001 & 3536 & djibrila\_diane\_parallel\_001.nqo\_Nkoo
        & \multirow{2}{*}{Djibrila Diane}
        & \multirow{2}{*}{\shortstack{Short Phrases in Various Tenses}} \\
        1001 & 3487 & djibrila\_diane\_parallel\_001.eng\_Latn \\
        \hline
        148 & 1303 & djibrila\_diane\_parallel\_002.nqo\_Nkoo
        & \multirow{2}{*}{Djibrila Diane}
        & \multirow{2}{*}{\shortstack{Various Short Phrases}} \\
        148 & 1361 & djibrila\_diane\_parallel\_002.eng\_Latn \\
    \bottomrule
    \end{tabular}
    \end{adjustbox}
    \caption{nicolingua-0005's bilingual subsets in Nko (\emph{nqo\_Nkoo}) and  English (\emph{eng\_Latn})}
    \label{tab:nicolingua-0005-bilingual-corpora-nqo-eng}
\end{table*}

\begin{table*}
    \centering
    \begin{adjustbox}{max width=0.95\textwidth}
    \begin{tabular}{rrlll}
    \toprule
        lines & words & file & originator & description\\
    \midrule
        37894 & 40436 & baba\_mamadi\_diane\_parallel\_001.nqo\_Nkoo
        & \multirow{2}{*}{Baba Mamadi Diane}
        & \multirow{2}{*}{\shortstack{Nko-Francais Dictionary}} \\
        37894 & 41598 & baba\_mamadi\_diane\_parallel\_001.fra\_Latn \\
        \hline
        3604 & 39020 & nafadji\_sory\_conde\_parallel\_001.nqo\_Nkoo
        & \multirow{2}{*}{Nafadji Sory Conde}
        & \multirow{2}{*}{\shortstack{Various Short Phrases}} \\
        3604 & 35037 & nafadji\_sory\_conde\_parallel\_001.fra\_Latn \\
        \hline
        1141 & 22379 & nafadji\_sory\_conde\_parallel\_003.nqo\_Nkoo
        & \multirow{2}{*}{Nafadji Sory Conde}
        & \multirow{2}{*}{\shortstack{Segment from ``L'enfant Noir''}} \\
        1141 & 21049 & nafadji\_sory\_conde\_parallel\_003.fra\_Latn \\
        \hline
        2200 & 16091 & nafadji\_sory\_conde\_parallel\_002.nqo\_Nkoo
        & \multirow{2}{*}{Nafadji Sory Conde}
        & \multirow{2}{*}{\shortstack{Phrases related to Guinean \\Society and Sociology}} \\
        2200 & 15413 & nafadji\_sory\_conde\_parallel\_002.fra\_Latn \\
        \hline
        721 & 11863 & nafadji\_sory\_conde\_parallel\_004.nqo\_Nkoo
        & \multirow{2}{*}{Nafadji Sory Conde}
        & \multirow{2}{*}{\shortstack{Guinean Constitution}} \\
        721 & 11345 & nafadji\_sory\_conde\_parallel\_004.fra\_Latn \\
    \bottomrule
    \end{tabular}
    \end{adjustbox}
    \caption{nicolingua-0005's bilingual subsets in Nko (\emph{nqo\_Nkoo}) and  French (\emph{fra\_Latn})}
    \label{tab:nicolingua-0005-bilingual-corpora-nqo-fra}
\end{table*}

\begin{table*}
    \centering
    \begin{adjustbox}{max width=0.95\textwidth}
    \begin{tabular}{rrlll}
    \toprule
        lines & words & file & originator & description\\
    \midrule
        134000 & 2017158 & nafadji\_sory\_conde\_monolingual\_001.nqo\_Nkoo
        & \multirow{1}{*}{Nafadji Sory Conde}
        & \multirow{1}{*}{\shortstack{Various Books and News Papers}} \\
        44604 & 853464 & baba\_mamadi\_diane\_monolingual\_002.nqo\_Nkoo
        & \multirow{1}{*}{Baba Mamadi Diane}
        & \multirow{1}{*}{\shortstack{Various Books and Articles}} \\
        10195 & 420749 & baba\_mamadi\_diane\_monolingual\_001.nqo\_Nkoo
        & \multirow{1}{*}{Baba Mamadi Diane}
        & \multirow{1}{*}{\shortstack{Various Books and Articles}} \\
    \bottomrule
    \end{tabular}
    \end{adjustbox}
    \caption{nicolingua-0005's monolingual subsets in Nko (\emph{nqo\_Nkoo})}
    \label{tab:nicolingua-0005-monolingual-corpora}
\end{table*}

% \clearpage

\section{Datasheet Questionnaire for \emph{nicolingua-0005} }
\label{sec::datasheet-questionnaire}

\subsection{Motivation}
\label{sec::datasheet-questionnaire-start}

\subsubsection{\textcolor{blue}{Who created the dataset(e.g., which team, research group) and on behalf of which entity (e.g. company, institution, organization)?}}
\emph{nicolingua-0005} was curated by Moussa Doumbouya (Stanford University). Its constituent corpora were provided by the following members of Nko USA Inc: Baba Mamadi Diane, Solo Farabado Cisse, Djibrila Diane, Nafadji Sory Conde, Kalo Mory Diane.

\subsubsection{\textcolor{blue}{Did they fund it themselves? If there is an associated grant, please provide the name of the grantor and the grant name and number.}}
Nko community members voluntarily composed the included corpora.

\subsubsection{\textcolor{blue}{For what purpose was the data set created? Was there a specific task in mind? If so, please specify the result type ( e.g. unit ) to be expected.}}
Some included corpora were composed specifically for the development of MT systems while others were originally created for educational purposes. See Appendix \ref{sec::nicolingua-0005-details} for details.

\subsubsection{\textcolor{blue}{Could any of these uses, or their results, interfere with human will or communicate a false reality?}}
Not to the best of our knowledge.

\subsubsection{\textcolor{blue}{What is the antiquity of the file? Provide, please, the current date.}}
July 19 2023.

\subsubsection{\textcolor{blue}{Has there been any monetary profit from the creation of this dataset?}}

No.

% cdm: I'm not sure why space is going wonky before this section title, but for the moment, fixing it manually:
\medskip

\subsection{Composition}

\subsubsection{\textcolor{blue}{Is there any synthetic data in the dataset? If so, in what percentage?}}
The corpus doesn't contain any synthetic data.

\subsubsection{\textcolor{blue}{Are there multiple types of instances or is there just one type? Please specify the type(s), e.g. Raw data, preprocessed, symbolic.}}
The corpus contains monolingual and parallel text corpora.

\subsubsection{\textcolor{blue}{What do the instances (of each  type, if appropriate) that comprise the data set represent? (e.g. documents, photos, people, countries).}}
The instances represent segments of text in Nko, English, and French.

\subsubsection{\textcolor{blue}{How many instances (of each type, if appropriate) are there in total?}}
See Tables \ref{tab:summary-of-nicolingua-0005}, \ref{tab:nicolingua-0005-trilingual-corpora}, \ref{tab:nicolingua-0005-bilingual-corpora-nqo-eng}, \ref{tab:nicolingua-0005-bilingual-corpora-nqo-fra}  and \ref{tab:nicolingua-0005-monolingual-corpora}

\subsubsection{\textcolor{blue}{Does the dataset contain all possible instances or is it just a sample of a larger set? i.e. Is the dataset different than an original one due to the preprocessing process? In case this dataset is a subset of another one, is the original dataset available?}}
This dataset is a collection of corpora from various sources.
Some sources were integrally sampled (e.g. quran), while other sources were composed by individual translators.

\subsubsection{\textcolor{blue}{Is there a label or a target associated with each of the instances? If so, please provide a description.}}
The multilingual subsets of the corpora are matching segments of text in multiple languages.

\subsubsection{\textcolor{blue}{What is the format of the data? e.g. .json, .xml, .csv .}}
The files are text files encoded in UTF-8 that have the following extensions matching the iso standard code of the language and writing system they contain:
.nqo\_Nkoo, .eng\_Latn, .fra\_Latn.

\subsubsection{\textcolor{blue}{Is any information missing from individual instances? If so, please provide a description, explaining why this information is missing (e.g. because it was unavailable). This does not include intentionally removed information, but might include, e.g. redacted text.}}
There is no missing information to report.

\subsubsection{\textcolor{blue}{Are there any errors, sources of noise, or redundancies in the dataset? If so, please provide a description. Do not include missing information here.}}
The sentences were benevolently translated by various individuals. A minimal quality control process was adopted during the curation phase. The data may contain some errors.
The corpus baba\_mamadi\_diane\_parallel\_003 was created by sampling Quran phrases from baba\_mamadi\_diane\_parallel\_002.
Some parallel Nko segments may be repeated in the monolingual Nko corpora.

\subsubsection{\textcolor{blue}{Is there any verification that guarantees there is not institutionalization of unfair biases? Both regarding the dataset itself and the potential algorithms that could use it.}}
no.

\subsubsection{\textcolor{blue}{Are there recommended data splits, e.g. training, development/validation, testing? If so, please provide a description of these splits explaining the rationale behind them.}}
The corpora are intended to be used to train natural language processing algorithms.

\subsubsection{\textcolor{blue}{Is the dataset self-contained, or does it link to or otherwise rely on external resources? e.g., websites, tweets, other datasets. If it links to or relies on external resources, a) Are there any guarantees that they will exist, and remain constant over time? b) Are there ofﬁcial archival versions of the complete dataset? i.e. including the external resources as they existed at the time the dataset was created. c) Are there any restrictions (e.g. licenses, fees) associated with any of the external resources that might apply to a future user? Please provide descriptions of all external resources and any restrictions associated with them, as well as links or other access points, if appropriate.}}
\emph{nicolingua-0005} is self-contained.

\subsubsection{\textcolor{blue}{Does the dataset contain data that might be considered conﬁdential? e.g. data that is protected by legal privilege or by doctor-patient conﬁdentiality, data that includes the content of individuals non-public communications. If so, please provide a description.}}
Not to the best of our knowledge.

\subsubsection{\textcolor{blue}{Does the dataset contain data that, if viewed directly, might be offensive, insulting, threatening, or might otherwise cause anxiety? If so, please describe why.}}
Not to the best of our knowledge.
Notes: (1) \emph{nicolingua-0005} contains religious text that some people may find offensive or threatening.
(2) Some words contained in \emph{nicolingua-0005}, such as the name of certain human body parts included in the Nko-Francais dictionary, may be considered vulgar or offensive.

\subsubsection{\textcolor{blue}{Does the dataset relate to people? If so, please specify a) Whether the dataset identifies subpopulations or not. b) Whether the dataset identifies indivual people or not. c) Whether it contains information that could vulnerate any individuals or their rights. c) Any other verified information on the topic that can be provided.}}
The data includes news articles that may reference specific people and people groups.
The data also includes literature relating to West African people and people groups and their history.

\subsubsection{\textcolor{blue}{Does the dataset cover included languages equally?}}
No. The sizes of various parallel and monolingual subsets have been specified in Table \ref{tab:summary-of-nicolingua-0005}.

\subsubsection{\textcolor{blue}{Is there any evidence that the data may be somehow biased? i.e. towards gender, ethics, beliefs.}}
The data includes religious texts, articles, and books that may reflect various types of biases.
The data may contain biases inherent in historical and current Manding culture such as work organization between men and women, young and old people.
Nko doesn't have masculine vs. feminine noun classes. Therefore genders are not distinguished in Nko nouns and pronouns, which may reduce the potential for gender-based bias.

\subsubsection{\textcolor{blue}{Is the data made up of formal text, informal text or both equitably?}}
The data mostly contains formal text.

\subsubsection{\textcolor{blue}{Does the data contain incorrect language expressions on purpose? Does it contain slang terms? If that’s the case, please provide which instances of the data correspond to these.}}
Not to the best of our knowledge.
The dataset may contain unintentional errors.

% cdm: I'm not sure why space is going wonky before this section title, but for the moment, fixing it manually:
\medskip

\subsection{Collection Process}

\subsubsection{\textcolor{blue}{Where was the data collected at? Please include as much detail; i.e. country, city, community, entity and so on. 
}}
Most data was collected in Conakry, Guinea, and Banakoro, Guinea. Some contributors also worked in Bamako, Mali (Solo F Cisse, Baba M Diane), Egypt (Baba M Diane) and USA (Djibrila Diane) while collecting the datasets.

\subsubsection{\textcolor{blue}{If the dataset is a sample from a larger set, what was the sampling strategy? i.e. deterministic, probabilistic with specific sampling probabilities.}}
N/A

\subsubsection{\textcolor{blue}{Are there any guarantees that the acquisition of the data did not violate any law or anyone's rights?}}
Not to the best of our knowledge.

\subsubsection{\textcolor{blue}{Are there any guarantees that prove the data is reliable?}}
No.

\subsubsection{\textcolor{blue}{Did the collection process involve the participation of individual people? If so, please report any information available regarding the following questions: Was the data collected from people directly? Did all the involved parts give their explicit consent? Is there any mechanism available to revoke this consent in the future, if desired?}}
The dataset authors are authors of this paper. They gave their explicit consent.

\subsubsection{\textcolor{blue}{Has an analysis of the potential impact of the dataset and its use on data subjects been conducted? i.e. a data protection impact analysis. If so, please provide a description of this analysis, including the outcomes, as well as a link or other access point to any supporting documentation.}}
No.

\subsubsection{\textcolor{blue}{Were any ethical review processes conducted?}}
No.

\subsubsection{\textcolor{blue}{Does the data come from a single source or is it the result of a combination of data coming from different sources? In any case, please provide references.}}
The data was curated from a combination of different sources.

\subsubsection{\textcolor{blue}{If the same content was to be collected from a different source, would it be similar?}}
Not Applicable.

\subsubsection{\textcolor{blue}{Please specify any other information regarding the collection process. i.e. Who collected the data, whether they were compensated or not, what mechanisms were used. Please, only include if verified.}}

% cdm: I'm not sure why space is going wonky before this section title, but for the moment, fixing it manually:
\medskip

\subsection{Preprocessing/Cleaning/Labelling}

\subsubsection{\textcolor{blue}{Please specify any information regarding the preprocessing that you may know (e.g. the person who created the dataset has somehow explained it) or be able to find (e.g. there exists and informational site). Please, only include if verified. i.e. Was there any mechanism applied to obtain a neutral language? Were all instances preprocessed the same way?}}
The data was normalized with Unicode normalization form NFC: Canonical Decomposition
followed by Canonical Composition.
Non-Nko characters were stripped from monolingual Nko text.
Extra punctuations were removed from some sources.
Some entries in Baba Mamadi Diane's Nko-Francais dictionary were expanded using regular expressions so that separate forms of the same words (e.g. gendered, plural) were repeated as separate entries.

% cdm: I'm not sure why space is going wonky before this section title, but for the moment, fixing it manually:
\medskip

\subsection{Uses}

\subsubsection{\textcolor{blue}{Has the dataset been used already? If so, please provide a description.}}
The data was used to build baseline neural machine translation algorithms for Nko. See Section \ref{sec:baseline-nmt-experiments}.

\subsubsection{\textcolor{blue}{Is there a repository that links to any or all papers or systems that use this dataset? If so, please provide a link or any other access point.}}

\url{https://github.com/mdoumbouya/nicolingua-0005-nqo-nmt-resources}
\url{https://github.com/mdoumbouya/nicolingua-0005-nqo-nmt-resources}

\subsubsection{\textcolor{blue}{What (other) tasks could the dataset be used for? Please include your own intentions, if any.}}
Any natural language processing tasks including language modeling and machine translation.

\subsubsection{\textcolor{blue}{Are there tasks for which the dataset should not be used? If so, please provide a description.}}
Not to the best of our knowledge.

\subsection{Distribution}

\subsubsection{\textcolor{blue}{Please specify the source where you got the dataset from.}}
The datasets came from the following individuals:

\subsubsection{\textcolor{blue}{When was the dataset first released?}}
July 19 2023.

\subsubsection{\textcolor{blue}{Are there any restrictions regarding the distribution and/or usage of this data in any particular geographic regions?}}
No.

\subsubsection{\textcolor{blue}{Is the dataset distributed under a copyright or other intellectual property (IP) license? And/or under applicable terms of use (ToU)? Please cite a verified source.}}

The dataset is openly available under the 
Attribution-ShareAlike 4.0 International (CC BY-SA 4.0) license.

\subsection{Maintenance}
\label{sec::datasheet-questionnaire-end}

\subsubsection{\textcolor{blue}{Is there any verified manner of contacting the creator of the dataset?}}
The authors of this paper can be contacted via email.

\subsubsection{\textcolor{blue}{Specify any limitations there might be to contributing to the dataset. i.e. Can anyone contribute to it? Can someone do it at all?}}
The dataset is openly available under the 
Attribution-ShareAlike 4.0 International (CC BY-SA 4.0) license.

\subsubsection{\textcolor{blue}{Has any erratum been notified?}}
No.

\subsubsection{\textcolor{blue}{Is there any verified information on whether the dataset will be updated in any form in the future? Is someone in charge of checking if any of the data has become irrelevant throughout time? If so, will it be removed or labeled somehow?}}
The dataset will be maintained on GitHub. Any updates will be made available in the same GitHub repository.

\subsubsection{\textcolor{blue}{Is there any available log about the changes performed previously in the dataset?}}
Any future modifications will be tracked in GitHub's version control.

\subsubsection{\textcolor{blue}{Could changes to current legislation end the right-of-use of the dataset?}}
Not to the best of our knowledge.

\subsubsection{\textcolor{blue}{Are there any lifelong learning updates, such as vocabulary enrichment, automatically developed?}}
No.

\clearpage
%%%%%%%%%%%%%%%
% NMT Appendices
%%%%%%%%%%%%%%%

\section{Train, Valid and Test Subset Details}
Details on the training, validation, and test subset composition for each model appear on the following page.

% \section{Learning Curves}

\begin{table*}[tbh]
    \centering
    \footnotesize
    \begin{tabular}{@{}rrllccc@{}}
        \multicolumn{7}{c}{TRAIN} \\
        lines & words & file & 200 & 201 & 202-9 & ~ \\
    \midrule
        6193 & 148442 & common-parallel-corpora/multitext-nllb-seed/bam\_Latn & ~ & \checkmark & \checkmark \\
        6193 & 136157 & cpc/multitext-nllb-seed/eng\_Latn & \checkmark & \checkmark & \checkmark \\
        6193 & 184138 & cpc/multitext-nllb-seed/nqo\_Nkoo & \checkmark & \checkmark & \checkmark \\
    \hline
        6236 & 151323 & nicolingua-0005/baba\_mamadi\_diane\_parallel\_002.eng\_Latn & \checkmark & \checkmark & \checkmark \\
        6236 & 171085 & nicolingua-0005/baba\_mamadi\_diane\_parallel\_002.fra\_Latn & ~ & \checkmark & \checkmark \\
        6236 & 175382 & nicolingua-0005/baba\_mamadi\_diane\_parallel\_002.nqo\_Nkoo & \checkmark & \checkmark & \checkmark \\
    \hline
        7001 & 17558 & nicolingua-0005/kalo\_mory\_diane\_parallel\_001.eng\_Latn & \checkmark & \checkmark & \checkmark \\
        7001 & 21593 & nicolingua-0005/kalo\_mory\_diane\_parallel\_001.fra\_Latn & ~ & \checkmark & \checkmark \\
        7001 & 28626 & nicolingua-0005/kalo\_mory\_diane\_parallel\_001.nqo\_Nkoo & \checkmark & \checkmark & \checkmark \\
    \hline
        4001 & 12891 & nicolingua-0005/kalo\_mory\_diane\_parallel\_003.eng\_Latn & \checkmark & \checkmark & \checkmark \\
        4001 & 15050 & nicolingua-0005/kalo\_mory\_diane\_parallel\_003.fra\_Latn & ~ & \checkmark & \checkmark \\
        4001 & 18864 & nicolingua-0005/kalo\_mory\_diane\_parallel\_003.nqo\_Nkoo & \checkmark & \checkmark & \checkmark \\
    \hline
        3999 & 12237 & nicolingua-0005/kalo\_mory\_diane\_parallel\_002.eng\_Latn & \checkmark & \checkmark & \checkmark \\
        3999 & 14495 & nicolingua-0005/kalo\_mory\_diane\_parallel\_002.fra\_Latn & ~ & \checkmark & \checkmark \\
        3999 & 17903 & nicolingua-0005/kalo\_mory\_diane\_parallel\_002.nqo\_Nkoo & \checkmark & \checkmark & \checkmark \\
    \hline
        3052 & 9615 & nicolingua-0005/solo\_farabado\_cisse\_parallel\_002.eng\_Latn & \checkmark & \checkmark & \checkmark \\
        3052 & 11308 & nicolingua-0005/solo\_farabado\_cisse\_parallel\_002.fra\_Latn & ~ & \checkmark & \checkmark \\
        3052 & 13420 & nicolingua-0005/solo\_farabado\_cisse\_parallel\_002.nqo\_Nkoo & \checkmark & \checkmark & \checkmark \\
    \hline
        1559 & 2382 & nicolingua-0005/solo\_farabado\_cisse\_parallel\_001.eng\_Latn & \checkmark & \checkmark & \checkmark \\
        1559 & 2338 & nicolingua-0005/solo\_farabado\_cisse\_parallel\_001.fra\_Latn & ~ & \checkmark & \checkmark \\
        1559 & 2739 & nicolingua-0005/solo\_farabado\_cisse\_parallel\_001.nqo\_Nkoo & \checkmark & \checkmark & \checkmark \\
    \hline
        21590 & 133369 & nicolingua-0005/baba\_mamadi\_diane\_parallel\_003.eng\_Latn & \checkmark & \checkmark & \checkmark \\
        21590 & 154238 & nicolingua-0005/baba\_mamadi\_diane\_parallel\_003.nqo\_Nkoo & \checkmark & \checkmark & \checkmark \\
    \hline
        36211 & 72612 & nicolingua-0005/baba\_mamadi\_diane\_parallel\_004.eng\_Latn & \checkmark & \checkmark & \checkmark \\
        36211 & 119536 & nicolingua-0005/baba\_mamadi\_diane\_parallel\_004.nqo\_Nkoo & \checkmark & \checkmark & \checkmark \\
    \hline
        1001 & 3487 & nicolingua-0005/djibrila\_diane\_parallel\_001.eng\_Latn & \checkmark & \checkmark & \checkmark \\
        1001 & 3536 & nicolingua-0005/djibrila\_diane\_parallel\_001.nqo\_Nkoo & \checkmark & \checkmark & \checkmark \\
    \hline
        148 & 1361 & nicolingua-0005/djibrila\_diane\_parallel\_002.eng\_Latn & \checkmark & \checkmark & \checkmark \\
        148 & 1303 & nicolingua-0005/djibrila\_diane\_parallel\_002.nqo\_Nkoo & \checkmark & \checkmark & \checkmark \\
    \hline
        492 & 4122 & nicolingua-0005/djibrila\_diane\_parallel\_003.eng\_Latn & \checkmark & \checkmark & \checkmark \\
        492 & 4666 & nicolingua-0005/djibrila\_diane\_parallel\_003.nqo\_Nkoo & \checkmark & \checkmark & \checkmark \\
    \hline
        37894 & 41598 & nicolingua-0005/baba\_mamadi\_diane\_parallel\_001.fra\_Latn & ~ & \checkmark & \checkmark \\
        37894 & 40436 & nicolingua-0005/baba\_mamadi\_diane\_parallel\_001.nqo\_Nkoo & ~ & \checkmark & \checkmark \\
    \hline
        3604 & 35037 & nicolingua-0005/nafadji\_sory\_conde\_parallel\_001.fra\_Latn & ~ & \checkmark & \checkmark \\
        3604 & 39020 & nicolingua-0005/nafadji\_sory\_conde\_parallel\_001.nqo\_Nkoo & ~ & \checkmark & \checkmark \\
    \hline
        2200 & 15413 & nicolingua-0005/nafadji\_sory\_conde\_parallel\_002.fra\_Latn & ~ & \checkmark & \checkmark \\
        2200 & 16091 & nicolingua-0005/nafadji\_sory\_conde\_parallel\_002.nqo\_Nkoo & ~ & \checkmark & \checkmark \\
    \hline
        1141 & 21049 & nicolingua-0005/nafadji\_sory\_conde\_parallel\_003.fra\_Latn & ~ & \checkmark & \checkmark \\
        1141 & 22379 & nicolingua-0005/nafadji\_sory\_conde\_parallel\_003.nqo\_Nkoo & ~ & \checkmark & \checkmark \\
    \hline
        721 & 11345 & nicolingua-0005/nafadji\_sory\_conde\_parallel\_004.fra\_Latn & ~ & \checkmark & \checkmark \\
        721 & 11863 & nicolingua-0005/nafadji\_sory\_conde\_parallel\_004.nqo\_Nkoo & ~ & \checkmark & \checkmark \\
    \hline
        134000 & 2017158 & nicolingua-0005/nafadji\_sory\_conde\_monolingual\_001.nqo\_Nkoo & ~ & ~ & \checkmark \\
        10195 & 420749 & nicolingua-0005/baba\_mamadi\_diane\_monolingual\_001.nqo\_Nkoo & ~ & ~ & \checkmark \\
        44604 & 853464 & nicolingua-0005/baba\_mamadi\_diane\_monolingual\_002.nqo\_Nkoo & ~ & ~ & \checkmark \\
        \bottomrule
    \end{tabular}
    \caption{Data files included in the training set of each model family}
    \label{tab:data-details-train}
\end{table*}

\begin{table*}[tbh]
    \centering
    \footnotesize
    \begin{tabular}{@{}llllccc@{}}
        \multicolumn{7}{c}{VALID} \\
        lines & words & file & 200 & 201 & 202-9 & ~ \\
    \midrule
        997 & 21565 & common-parallel-corpora/flores-200-dev/bam\_Latn.dev & ~ & \checkmark & \checkmark \\
        997 & 20954 & common-parallel-corpora/flores-200-dev/eng\_Latn.dev & \checkmark & \checkmark & \checkmark \\
        997 & 23957 & common-parallel-corpora/flores-200-dev/fra\_Latn.dev & ~ & \checkmark & \checkmark \\
        997 & 27361 & common-parallel-corpora/flores-200-dev/nqo\_Nkoo.dev & \checkmark & \checkmark & \checkmark \\
    \\
        \multicolumn{7}{c}{TEST} \\
        lines & words & file & 200 & 201 & 202-9 & ~ \\
    \midrule
        1012 & 22565 & common-parallel-corpora/flores-200-devtest/bam\_Latn.devtest & ~ & \checkmark & \checkmark \\
        1012 & 21901 & common-parallel-corpora/flores-200-devtest/eng\_Latn.devtest & \checkmark & \checkmark & \checkmark \\
        1012 & 25319 & common-parallel-corpora/flores-200-devtest/fra\_Latn.devtest & ~ & \checkmark & \checkmark \\
        1012 & 29503 & common-parallel-corpora/flores-200-devtest/nqo\_Nkoo.devtest & \checkmark & \checkmark & \checkmark \\
    \bottomrule
    \end{tabular}
    \caption{Data files included in the validation and test sets of each model family}
    \label{tab:data-details-valid-test}
\end{table*}

\clearpage

\section{Examples of Translations}
\label{sec:generation-sensitivity-examples}
Examples of generations highlighting the sensitivity our ouf baseline NMT system to punctuation and case appear on the following page.

\begin{figure*}[ht]
    \centering
    \includegraphics[width=0.7\textwidth]{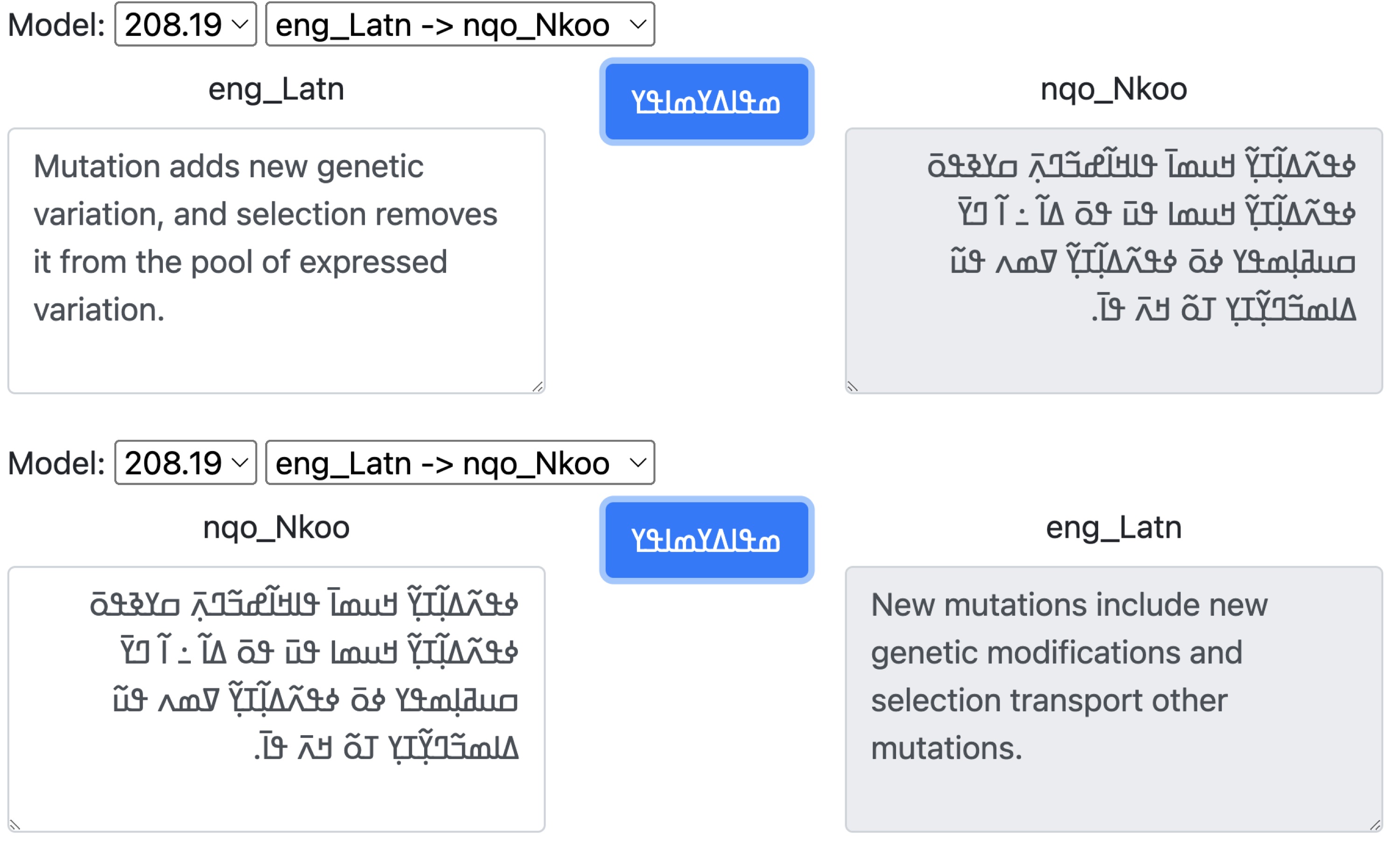}
    \\
    \rule[1ex]{8cm}{0.9pt}
    \\
    \includegraphics[width=0.7\textwidth]{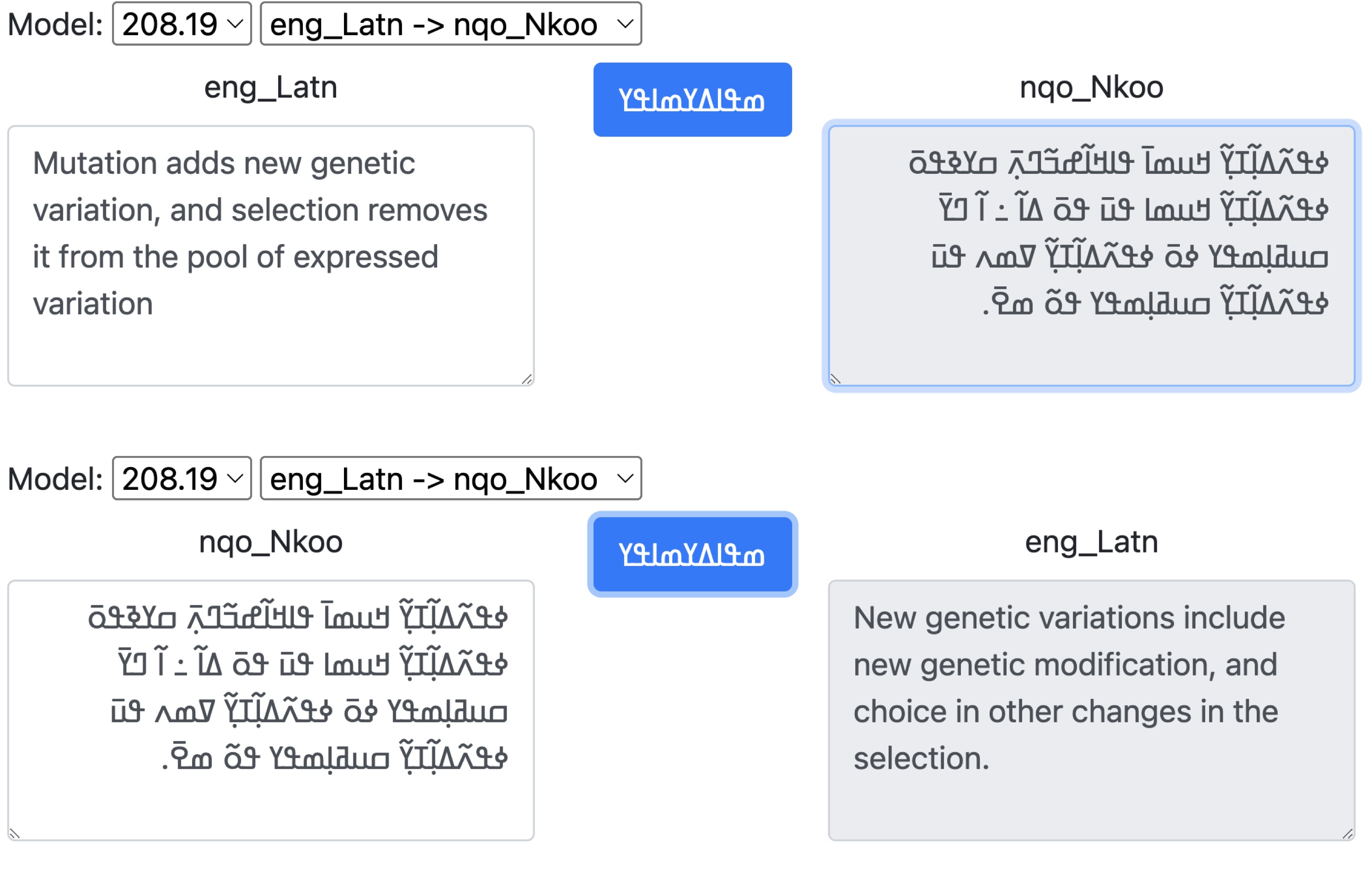}
    \\
    \rule[1ex]{8cm}{0.9pt}
    \\
    \includegraphics[width=0.7\textwidth]{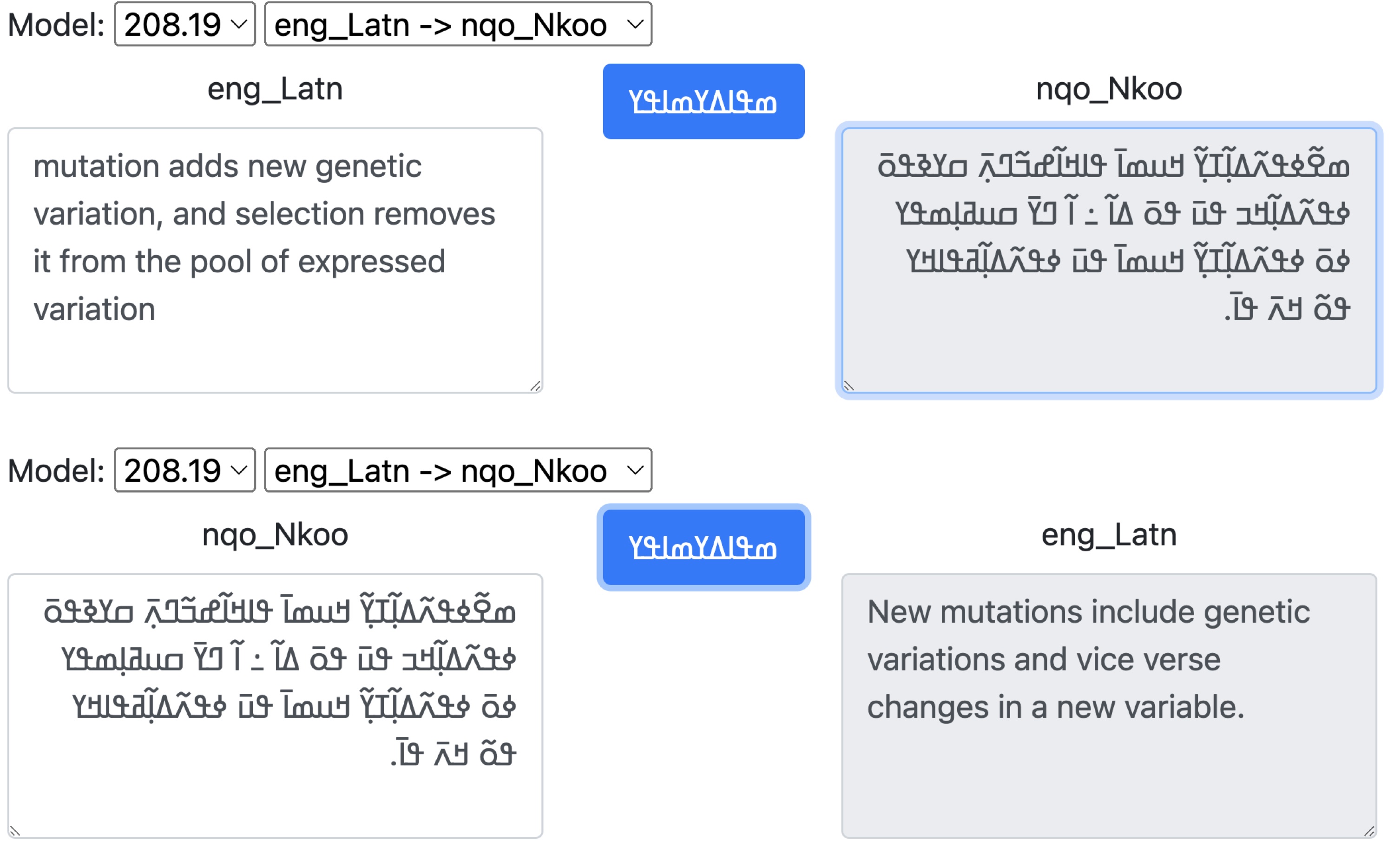}
    \caption{A sentence from the FLoRes-200-devtest corpus translated from English to Nko and back-translated to English using model 208.19. The three examples highlight the sensitivity of our baseline system to punctuation and case. Top: original sentence; Middle: removed final period; Bottom: removed initial capitalization and final period.}
    \label{fig:nko-nmt-system-sensitivity-to-punctuation-and-case}
\end{figure*}

\end{document}